\documentstyle[11pt,times,fullpage,myapa,epsfig,here]{article}

\newcommand{\setstretch}[1]{\relax}
\newenvironment{singlespace}{\relax}{\relax}

\makeatletter
\newcommand{\mysection}[1]{%
        \pagebreak[3]%
        \vspace{\baselineskip}
        \begin{center}\large\bf\uppercase{#1}\end{center}
        \nopagebreak\par}
\newcommand{\mysubsection}[1]{%
        \vspace{1.2\baselineskip}
        \par\noindent
        {\large\bf #1}%
        \nopagebreak\par}
\newcommand{\mysubsubsection}[1]{%
        \vspace{.8\baselineskip}
        \par{\bf #1}.~~~%
        \nopagebreak}

\date{{\it To appear in {\sc language and speech} 41(3-4): 439-487. Special Issue on Prosody and Conversation, 1998}}

\renewcommand{\section}{\@ifstar{\mysection}{\mysection}}
\renewcommand{\subsection}{\@ifstar{\mysubsection}{\mysubsection}}
\renewcommand{\subsubsection}{\@ifstar{\mysubsubsection}{\mysubsubsection}}

\makeatother

\renewcommand{\cite}[1]{\shortcite{#1}}
\newcommand{\namecite}[1]{\shortciteA{#1}}
\newcommand{\opencite}[1]{\shortciteNP{#1}}

\newcommand{\includetree}[1]{\strut\psfig{file=#1,angle=270,width=7in,height=5in}}
\newcommand{\includebargraph}[1]{\strut\psfig{file=#1,angle=90,width=5.8in,height=2.8in}}

\newcommand{\includeLMbargraph}[1]{\strut\psfig{file=#1,width=5.5in,height=8in}}

\newcommand{\PB}[1]{\parbox[t]{4.0in}{%
        \setlength{\baselineskip}{0.7\baselineskip} #1}}

\title{Can Prosody Aid the Automatic Classification of Dialog Acts \\
      in Conversational Speech?}

\author{Elizabeth~Shriberg, SRI International, Menlo Park, CA, U.S.A.\\
Rebecca~Bates, Boston University, Boston, MA, U.S.A.\\
Andreas~Stolcke, SRI International, Menlo Park, CA, U.S.A.\\
Paul~Taylor, University of Edinburgh, Edinburgh, U.K.\\
Daniel~Jurafsky, University of Colorado, Boulder, CO, U.S.A.\\
Klaus~Ries, Carnegie Mellon University, Pittsburgh, PA, U.S.A.\\
Noah~Coccaro, University of Colorado, Boulder, CO, U.S.A.\\
Rachel~Martin, Johns Hopkins University, Baltimore, MD, U.S.A.\\
Marie~Meteer, BBN Systems and Technologies, Cambridge, MA, U.S.A.\\
Carol~Van Ess-Dykema, U.S.~Department of Defense, Ft.~Meade, MD, U.S.A.}

\begin{document}

\setstretch{1.0}

\maketitle

\noindent{\bf Running Head:} Prosodic classification of dialog acts

\begin{singlespace}
\noindent{\bf Acknowledgments:}

This work was funded by the sponsors of the 1997 Workshop on Innovative
Techniques in Large Vocabulary Conversational Speech Recognition at the
Center for Speech and Language Processing at Johns Hopkins University,
and by the National Science Foundation through grants IRI-9619921 and
IRI-9314967 to Elizabeth Shriberg and IRI-970406 to Daniel Jurafsky. We
thank Fred Jelinek and Kimberly Shiring of the JHU Workshop for
supporting the project, and workshop participants Joe Picone, Bill Byrne,
and Harriet Nock for assistance with data resources and recognizer
software.  We are grateful to Susann LuperFoy, Nigel Ward, James Allen,
Julia Hirschberg, and Marilyn Walker for advice on the design of the
SWBD-DAMSL tag-set, to Mitch Weintraub for the SNR measurement software,
to Nelson Morgan, Eric Fosler-Lussier, and Nikki Mirghafori for making
the enrate software available, to James Hieronymus for discussion of
prosodic features, and to Julia Hirschberg and two anonymous reviewers
for helpful comments.
Special thanks go to the Boulder graduate students for dialog act
labeling: Debra Biasca (project manager), Traci Curl, Marion Bond, Anu
Erringer, Michelle Gregory, Lori Heintzelman, Taimi Metzler, and Amma
Oduro; and to the Edinburgh intonation labelers: Helen Wright, Kurt
Dusterhoff, Rob Clark, Cassie Mayo, and Matthew Bull. The views and
conclusions contained in this document are those of the authors and
should not be interpreted as representing official policies of the
funding agencies.

\end{singlespace}

\begin{singlespace}
\noindent{\bf Corresponding Author:}
\begin{quote}
Elizabeth Shriberg\\
SRI International\\
333 Ravenswood Ave.\\
Menlo Park, CA 94025\\
Tel: +1-650-859-3798\\
FAX: +1-650-859-5984     
\end{quote}
\end{singlespace}

\newpage

\setstretch{1.8}

\section{Abstract}

Identifying whether an utterance is a statement, question, greeting,
and so forth is integral to effective automatic understanding of
natural dialog. Little is known, however, about how such dialog acts
(DAs) can be automatically classified in truly natural conversation.
This study asks whether current approaches, which use mainly word
information, could be improved by adding prosodic information.

The study is based on more than 1000 conversations from the
Switchboard corpus.  DAs were hand-annotated, and prosodic features
(duration, pause, F0, energy, and speaking rate) were automatically
extracted for each DA\@.  In training, decision trees based on these
features were inferred; trees were then applied to unseen test data to
evaluate performance.  Performance was evaluated for prosody models
alone, and after combining the prosody models with word
information---either from true words or from the output of an
automatic speech recognizer.

For an overall classification task, as well as three subtasks, prosody
made significant contributions to classification.  Feature-specific
analyses further revealed that although canonical features (such as F0
for questions) were important, less obvious features could compensate
if canonical features were removed. Finally, in each task, integrating
the prosodic model with a DA-specific statistical language model
improved performance over that of the language model alone, especially
for the case of recognized words. Results suggest that DAs are
redundantly marked in natural conversation, and that a variety of
automatically extractable prosodic features could aid dialog
processing in speech applications.

{\bf Keywords:}
automatic dialog act classification, prosody, discourse modeling,
speech understanding, spontaneous speech recognition.

\newpage

\section{Introduction}

\subsection{Why Model Dialog?}

Identifying whether an utterance is a statement, question, greeting, and
so forth is integral to understanding and producing natural dialog.
Human listeners easily discriminate such dialog acts (DAs) in everyday
conversation, responding in systematic ways to achieve the mutual goals
of the participants \cite{Clark:96,Levelt:89}.  Little is known,
however, about how to build a fully automatic system that can
successfully identify DAs occurring in natural conversation.

At first blush, such a goal may appear misguided, because most current
computer dialog systems are designed for human-computer interactions
in specific domains.  Studying unconstrained human-human dialogs would
seem to make the problem more difficult than necessary, since
task-oriented dialog (whether human-human or human-computer) is by
definition more constrained and hence easier to process.
Nevertheless, for many other applications, as well as for basic
research in dialog, developing DA classifiers for conversational
speech is clearly an important goal. For example, optimal automatic
summarization and segmentation of natural conversations (such as
meetings or interviews) for archival and retrieval purposes requires
not only knowing the string of words spoken, but also who asked
questions, who answered them, whether answers were agreements or
disagreements, and so forth. Another motivation for speech technology
is to improve word recognition. Because dialog is highly conventional,
different DAs tend to involve different word patterns or phrases.
Knowledge about the likely DA of an utterance could therefore be
applied to constrain word hypotheses in a speech recognizer.  Modeling
of DAs from human-human conversation can also guide the design of
better and more natural human-computer interfaces.  On the theoretical
side, information about properties of natural utterances provides
useful comparison data to check against descriptive models based on
contrived examples or speech produced under laboratory settings.
Automatic methods for classifying dialog acts could also be applied to
the problem of labeling large databases when hand-annotation is not
feasible, thereby providing data to further basic research.

\subsection{Word-Based Approaches to Dialog Act Detection}

Automatic modeling of dialog has gained interest in recent years,
particularly in the domain of human-computer dialog applications.  One
line of work has focused on predicting the most probable next dialog act in
a conversation, using mainly information about the DA history or context
\cite{Yamaoka91,WosWai94,NagMor94,ReithingerMaier:95,BennacefEtAl:95,Kita96,Reithinger96}.  A
second, related line of research has focused on DA recognition and
classification, taking into account both the DA history and features of
the current DA itself
\cite{SuhmWaibel94,ReithingerKlesen:97,Chu-Carroll:aaai98,Samuel:aaai98}.
In all of these previous approaches, DA classification has relied
heavily on information that can be gleaned from words,
such as cue phrases and N-grams, or information that can be derived
from word sequences, such as syntactic form.

\subsection{Why Use Prosody?}

This work focuses on exploring another, relatively untapped
potential knowledge source for automatic DA classification: prosody.  By
prosody we mean information about temporal, pitch, and energy
characteristics of utterances that are independent of the words.  We
were interested in prosody for several reasons.  First, some DAs are
inherently ambiguous from word information alone. For example,
declarative questions (e.g., ``John is here?'') have the same word order
as statements, and hence when lexical and syntactic cues are consistent with that
of a statement, may be distinguishable as a question only via
prosody.  Second, in a real application, word recognition may not be
perfect. Indeed, state-of-the-art recognizers still show over 30\%
word error rate for large-vocabulary conversational speech.
Third, there are potential applications for which a full-fledged speech
recognizer may not be available or practical, and a less computationally
expensive, but somewhat less accurate method to track the structure of a 
dialog is acceptable.
Fourth, an understanding of prosodic properties of different utterance
types can lead to more natural output from speech synthesis systems. 
And finally, it is of basic theoretical interest to descriptive accounts
in linguistics, as well as to psycholinguistic theories of sentence processing,
to understand how different DAs are signaled prosodically.

\subsection{Previous Studies of Prosody and Discourse}

The main context in which prosody has been explored specifically for the
purpose of dialog processing is in the area of discourse {\em
  segmentation}---both at the utterance level and at higher levels such
as the organization of utterances into turns and topics.  The
segmentation studies span both descriptive and computational fields, and
describe or attempt to detect utterance and topic boundaries using
various acoustic-prosodic features, including pitch range, intonational
contour, declination patterns, utterance duration, pre-boundary
lengthening phenomena, pause patterns, speaking rate, and energy
patterns. There has been increasing work in studying spontaneous
speech, in both human-human and human-machine dialog. In most
cases the features cuing the segments are coded by hand, but could
potentially be estimated by automatic means for speech applications
\cite{GroHir:92,NakajimaAllen:93,Ayers:94,Litman:95,HirschbergNakatani:96,KoopmansDonzel:96,BruceEtAl:97,Nakajima:97,Swerts:97,SweOst:97}.
Although much of the work on prosody and segmentation has been
descriptive, some recent studies have developed classifiers and tested
performance using a fully automatic detection paradigm. For example,
\namecite{HirschbergNakatani:aaai98} found that features derived from a
pitch tracker (F0, but also voicing and energy information) provide cues
to intonational phrase boundaries; such a system could be used as a
front end for audio browsing and playback.  Similarly, in experiments on
subsets of the German Verbmobil spontaneous speech corpus, prosodic
features (including features reflecting duration, pause, F0, and energy)
were found to improve segmentation performance (into DAs) over that given by a
language model alone \cite{MastEtAl:96,Warnke:97}. 
The Verbmobil work was in the
context of an overall system for automatically classifying DAs, but the
prosodic features were used only at the segmentation stage.

A second line of relevant previous work includes studies on the
automatic detection of pitch accents, phrase accents, and boundary tones
for speech technology.  It has become increasingly clear that a
transcribed word sequence does not provide enough information for speech
understanding, since the same sequence of words can have different
meanings depending, in part, on prosody.  The location and type of
accents and boundary tones can provide important cues for tasks such as
lexical or syntactic disambiguation, and can be used to rescore word
hypotheses and reduce syntactic or semantic search complexity
\cite{Waibel:88,VeilleuxOstendorf:darpa93,WightmanOstendorf:94,KKN+:eurospeech95,Kompe:97}.
These and many related studies model F0, energy, and duration patterns
to detect and classify accents and boundary tones; information on the
location and type of prosodic events can then be used to assign or
constrain meaning, typically at the level of the utterance.  Such
information is relevant to dialog processing, since the locations of
major phrase boundaries delimit utterance units, and since tonal
information can specify pragmatic meaning in certain contexts (e.g.,
a rising final boundary tone suggests questions).  First
developed for formal speech, such approaches have also been
applied to spontaneous human-computer dialog, where the modeling problem
becomes more difficult as a result of less constrained speech styles.

Beyond the detection of accents, boundary tones, and discourse-relevant
segment boundaries, there has been only limited investigation into
automatic processing specifically to identify DAs in conversational
speech.  In one
approach, Taylor et al.~\citeyear{TaylorEtAl:97,TaylorEtAl:LS98}
used hidden Markov models (HMMs) to model
accents and boundary tones in different conversational ``moves'' in the
Maptask corpus \cite{carletta:95}, with the aim of applying
move-specific language models to improve speech recognition. The event
recognizer used ``tilt'' parameters \cite{TaylorBlack:94}, or F0,
amplitude, duration, and a feature capturing the shape (rise, fall, or
combination). As reported in many other studies of accent detection,
performance degraded sharply from speaker-dependent formal styles to
speaker-independent spontaneous speech (e.g., \opencite{OstendorfRoss:97}).
The automatic detection of moves was thus limited by somewhat low accent
detection accuracy (below 40\%); however, overall results suggested that
intonation can be a good predictor of move type.

In another study, \namecite{YoshimuraEtAl:96} aimed to automatically
identify utterances in human-machine dialog likely to contain emotional
content such as exclamations of puzzlement, self-talk, or other types of
paralinguistic information that the system would not be able to process.
The approach involved clustering utterances based on
vector-quantized F0 patterns and overall regression fits on the
contours. Patterns deviating from a typically relatively flat overall
slope were found to be likely to contain such paralinguistic content. 

Finally, researchers on the Verbmobil project
\cite{Kiessling:93,KKN+:eurospeech95}, following ideas of
\namecite{Noeth:91}, addressed an interesting case of ambiguity in
human-machine interaction in the context of a train-scheduling system.
Apparently, subjects often interrupt the announcement of train schedules
to repeat a specific departure or arrival time.  The repeat can serve
one of three functional roles: confirmation of understanding,
questioning of the time, or feedback that the user is still listening.
The tendency of users to interrupt in this manner is even more
pronounced when talking to an automatic system with synthesized speech
output, since the synthesis can often be difficult to comprehend.  To
aid in automatically identifying responses, Gaussian classifiers were
trained on F0 features similar to those mentioned in earlier work
\cite{Waibel:88,DalyZue:92}, including the slope of the regression line of
the whole contour and of the final portion, as well as utterance onset- and
offset-related values.
Similarly, \namecite{Terry94} used F0 information to distinguish
user queries from acknowledgments in a direction-giving system.
To this end, the shape of pitch contours was classified either
by a hand-written rule system, or a trained neural network.

\subsection{Current Study}

For the present work, we were interested in automatic methods that could
be applied to spontaneous human-human dialog, which is notoriously more
variable than read speech or most forms of human-computer dialog
\cite{DalyZue:92,Ayers:94,Blaauw:95}.  We also wanted to cover the full
set of dialog act labels observed, and thus needed to be able to define
the extraction and computation of all proposed features for all
utterances in the data. We took an exploratory approach, including a
large set of features from the different categories of prosodic features
used in the work on boundary and discourse described earlier.  However,
our constraints were somewhat different than in previous studies.

One important difference is that because we were interested in using
prosodic features in combination with a language model in speech
recognition, our features were designed to not rely on any word
information;
as explained later, this feature independence allows a probabilistic
combination of prosodic and word-based models.
A second major difference between our approach and work
based on hand-labeled prosodic annotations is that our features
needed to be automatically extractable from the signal.  This constraint was
practical rather than theoretical: it is currently not
feasible to automatically detect abstract events such as accents and
phrase boundaries reliably in spontaneous human-human dialog with
variable channel quality (such as in telephone speech).  Nevertheless, it
is also the case that we do not yet fully understand how abstract
categories characterize DAs in natural speech styles, and that an
understanding could be augmented by information about correlations
between DAs and other feature types. For example, even for DAs with
presumed canonical boundary tone indicators (such as the rising
intonation typical of questions), other features may additionally characterize
the DA.
For instance, descriptive analyses of Dutch question intonation 
have found that in addition to a final F0 rise, certain interrogatives
differ from declaratives in features located elsewhere, such as in
onset F0 and in overall pitch range \cite{Haan1EtAl:97,Haan2EtAl:97}.
Thus, we focussed on
global and rather simple features, and assumed no landmarks in our
utterances other than the start and end times.

Our investigation began as part of a larger project
\cite{JuretalASRU:97,JuretalTR:98,Stolcke:aaai98} on DA classification
in human-human telephone conversations, using three knowledge sources:
(1)~a dialog grammar (a statistical model of the sequencing of DAs in a
conversation), (2)~DA-specific language models (statistical models of
the word sequences associated with particular types of DAs), and
(3)~DA-specific prosodic models.  Results revealed that the modeling was
driven largely by DA priors (represented as unigram frequencies in the
dialog grammar) because of an extreme skew in the distribution of DAs in
the corpus---nearly 70\% of the utterances in the corpus studied were
either statements (declaratives) or brief backchannels (such as
``uh-huh'').
Because of the skew, it was difficult to assess the potential contribution
of features of the DAs themselves, including the prosodic features.
Thus, to better investigate
whether prosody can contribute to DA classification in
natural dialog, for this paper we eliminate additional knowledge
sources that could confound our results.  Analyses are conducted in a 
domain of uniform priors (all DAs are made equally likely). We also exclude
contextual information from the dialog grammar (such as the DA of the
previous utterance).  In this way, we hope to gain a better
understanding of the inherent prosodic properties of different DAs,
which can in turn help in the building of better integrated models for
natural speech corpora in general.

Our approach builds on a methodology previously developed for a different
task involving conversational speech \cite{ShrBatSto:97}. The
method is based on constructing a large database of automatically
extracted acoustic-prosodic features. In training, decision tree
classifiers are inferred from the features; the trees are then applied to
unseen data to evaluate performance and to study feature usage.

The analyses examine decision tree performance in four DA-classification tasks.
We begin with a task involving multiway classification of the DAs in our
corpus. We then examine three binary classification tasks found to be
problematic for
word-based classification: Question detection, Agreement
detection, and the detection of Incomplete Utterances.  For
each task, we train classifiers using  various subsets of features to gain an
understanding of the relative importance of different feature types. 
In addition, we integrate tree models with DA-specific language models
to explore the role of prosody when word information is also available,
from either a transcript or a speech recognizer.

\section{Method}

\subsection{Speech Data}

Our data were taken from the Switchboard corpus of human-human telephone
conversations on various topics \cite{Godfrey92}.  The original release
of this corpus contains roughly three million words from more than 2430
different conversations, each roughly 10 minutes in duration. The corpus
was collected at Texas Instruments and is distributed by the Linguistics
Data Consortium (LDC). A set of roughly 500 speakers representing all major
dialects of American English participated in the task in exchange for a
per-call remuneration. Speakers could participate as often as they
desired; many speakers participated multiple times.  Speakers were aware
that their speech was being recorded, but were informed only generally that
TI speech researchers were interested in the conversations.  Speakers
registered by choosing topics of interest (e.g., recycling, sports) from
a predetermined set, and by indicating times that they would be
available. They were automatically connected to another caller by a
``robot operator'' based on matching of registrants to topics and
available times. An advantage of this procedure is the absence of
experimenter bias. Conversations were therefore between strangers;
however, transcribers rated the majority of conversations as sounding
highly ``natural''. There were some clear advantages to using this
corpus for our work, including its size, the availability of
transcriptions, and sentence-level segmentations. But most important,
it was one of the only large English conversational-speech corpora
available at the time, for which we could obtain N-best word recognition
output from a state-of-the-art recognition system.

\subsection{Dialog Act Labeling}

\subsubsection{Labeling system}
We developed a DA labeling system for Switchboard, taking as a
starting point the DAMSL system \cite{CoreAllen:97} of DA labeling for
task-oriented dialog.  We adapted the DAMSL system to allow better
coverage for Switchboard, and also to create labels that provide
more information about the lexical and syntactic realization of DAs.
Certain classes in DAMSL were
never used, and conversely it was necessary to expand some of the
DAMSL classes to provide a variety of labels. 
The adapted system, ``SWBD-DAMSL'', is described in detail in
\namecite{Jurafsky:97-damsl}.

\begin{table}[H]
\begin{singlespace}
\caption{Seven Grouped Dialog Act Classes}
\label{tab:seven-DA-classes}

\begin{center}
\begin{tabular}{|l|c|l|}
\hline
       Type            & SWBD-DAMSL Tag          & Example \\
\hline
\hline
       Statements      & & \\
        \quad Description & sd        &{\em  Me,  I'm in the legal department}  \\
        \quad View/Opinion & sv       &{\em I think it's great} \\
\hline
       Questions       & & \\
        \quad Yes/No   & qy      & {\em Do you have to have any special training?}\\
        \quad Wh        & qw     & {\em Well, how old are you?} \\
        \quad Declarative & qy\^{}d, qw\^{}d    &{\em So you can afford to get a house?} \\
        \quad Open      & qo   & {\em How about you?}\\
\hline
       Backchannels    & b      & {\em Uh-huh} \\
\hline
       Incomplete Utterances & \%     & {\em So, -} \\
\hline
       Agreements      & aa     & {\em That's exactly it} \\
\hline
       Appreciations   & ba     & {\em I can imagine} \\
\hline
       Other           & all other &  (see Appendix A)    \\
\hline
\end{tabular}
\end{center}
\end{singlespace}
\end{table}

SWBD-DAMSL defines approximately 60 unique tags, many of which represent
orthogonal information about an utterance and hence can be combined.
The labelers made use of 220 of these combined tags, which we clustered
for our larger project into 42 classes \cite{JuretalTR:98}.  To simplify
analyses, the 42 classes were further grouped into
seven disjoint main classes, consisting of the frequently occurring
classes plus an ``Other'' class containing DAs each occurring less than
2\% of the time.  The groups are shown in
Table~\ref{tab:seven-DA-classes}.  The full set of DAs is listed in
Appendix~A, along with actual frequencies.
The full list is useful for getting a feel for the heterogeneity of the
``Other'' class.  Table~\ref{tab:swbd-example} shows three typical exchanges
found in the corpus, along with the kinds of annotations we had at our disposal.

\begin{table}[H]
\begin{singlespace}
\begin{center}
\caption{\label{tab:swbd-example} Example Exchanges in Switchboard.
 Utterance boundaries are indicated by ``/''; ``-/'' marks incomplete utterances.}
\small
\begin{tabular}{|cllll|}\hline
Speaker &\multicolumn{2}{c}{Dialog Act}&\multicolumn{2}{c|}{Utterance}\\ 
\hline
\hline
{\bf A}&\multicolumn{2}{l}{\bf Wh-Question} &\multicolumn{2}{l|}{What kind do you have now? /}\\
{\bf B}&\ ~~~~~~~~~~~&{\bf Statement-non-opinion} & ~~~ &{\em Uh, we have a, a Mazda
nine twenty nine and a Ford} \\
& & & & {\em  Crown Victoria and a little two seater CRX.} / \\
{\bf A}&\multicolumn{2}{l}{\bf Acknowledge-Answer} &\multicolumn{2}{l|}{Oh, okay. /}  \\
{\bf B}&&{\bf Statement-Opinion} &&{\em Uh, it's rather difficult to, to project what
kind of, uh,} -/ \\
{\bf A}&\multicolumn{2}{l}{\bf Statement-non-opinion}& \multicolumn{2}{l|}{We'd, look,
always look into, uh, consumer reports to see what kind} \\
& & &\multicolumn{2}{l|}{of, uh, report, or, uh, repair records that the
various cars have -/} \\
 \hline
 \hline
{\bf B}&& {\bf Abandoned}& &{\em So, uh,} -/\\
{\bf A}&\multicolumn{2}{l}{\bf Yes-No-Question}&\multicolumn{2}{l|}{And did
you find that you like the foreign cars better than the domestic? /} \\
{\bf B}&&{\bf Yes-Answer} &&{\em Uh, yeah.} /\\
\hline
\hline
{\bf B}&&{\bf Statement-non-opinion}&&{\em We've been extremely pleased with our Mazdas.} /\\
{\bf A}&\multicolumn{2}{l}{\bf Backchannel-Question}&\multicolumn{2}{l|}{Oh, really? /}    \\
{\bf B}&&{\bf Yes-Answer}&& {\em Yeah.} /\\
\hline
\end{tabular}
\end{center}
\end{singlespace}
\end{table}

For the Statement classes, independent analyses showed that the two
SWBD-DAMSL types of Statements, Descriptions and Opinions, were 
similar in their lexical and their prosodic features, although they did
show some differences in their distribution in the discourse, which
warrants their continued distinction in the labeling system. Since, as
explained in the Introduction, we do not use dialog grammar information
in this work, there is no reason not to group the two types together for
analysis.  For the Question category we grouped together the
main question types described by \namecite{Haan1EtAl:97,Haan2EtAl:97},
namely, Declarative Questions, Yes-No Questions, and Wh-Questions.

\subsubsection{Labeling procedure}
Since there was a large set of data to label, and limited time and labor
resources, we decided to have our main set of DA labels produced based
on the text transcripts alone.  Llabelers were given the
transcriptions of the full conversations, and thus could use contextual
information, as well as cues from standard punctuation (e.g., question
marks), but did not listen to the soundfiles. A similar approach
was used for the same reason in the work of \namecite{MastEtAl:96}. We
were aware, however, that labeling without listening is not without
problems. One concern is that certain DAs are inherently ambiguous from
transcripts alone. A commonly noted example is the distinction between
simple Backchannels, which acknowledge a contribution (e.g., ``uh-huh'')
and explicit Agreements (e.g., ``that's exactly it'').  There is
considerable lexical overlap between these two DAs, with emphatic
intonation conveying an Agreement (e.g., ``right'' versus ``right!'').
Emphasis of this sort was not marked by punctuation in the
transcriptions, and Backchannels were nearly four times as likely in our
corpus; thus, labelers when in doubt were instructed to mark an
ambiguous case as a Backchannel.  We therefore expected that some
percentage of our Backchannels were actually Agreements. In addition to
the known problem of Backchannel/Agreement ambiguities, we were
concerned about other possible mislabelings.  For example, rising
intonation could reveal that an utterance is a Declarative Question
rather than a Statement. Similarly, hesitant-sounding prosody could
indicate an Incomplete Utterance (from the point of view of the
speaker's intention), even if the utterance is potentially complete
based on words alone.

Such ambiguities are of particular concern for the analyses at hand,
which seek to determine the role of prosody in DA classification. If
some DAs are identifiable only when prosody is made available, then a
subset of our original labels will not only be {\em incorrect}, they
will also be {\em biased} toward the label cued by a language model.
This will make it difficult to determine the degree to which prosodic
cues can contribute to DA classification above and beyond the language
model cues.  We took two steps toward addressing these concerns within
the limits of our available resources.  First, we instructed our
labelers to flag any utterances that they felt were ambiguous from text
alone. In future work such utterances could be labeled after listening.
Given that this was not possible yet for all of the labeled data, we
chose to simply remove all flagged utterances for the present analyses.

Second, we conducted experiments to assess the loss incurred by
labeling with transcripts only. We asked one of the most experienced
of our original DA labelers%
\footnote{We thank Traci Curl for
  reannotating the data and for helpful discussions.} to reannotate
utterances after listening to the soundfiles.  So that the factor of
listening would not be confounded with that of inter-labeler agreement, all
conversations to be relabeled were taken from the set of conversations
that she had labeled originally.  In the interest of time, the
relabeling was done with the original labels available. Instructions
were to listen to all of the utterances, and take the time needed to
make any changes in which she felt the original labels were inconsistent
with what she heard.  This approach is not necessarily equivalent to
relabeling from scratch, since the labeler may be biased toward
retaining previous labels.  Nevertheless, it should reveal the types of
DAs for which listening is most important. This was the goal of a first
round (Round I) of relabeling, in which we did not give any
information about which DAs to pay attention to. The rate of changes for
the individual DA types, however, was assumed to be conservative here,
since the labeler had to divide her attention over all DA types.
Results are shown in the left column of Table~\ref{tab:relabeling}.

\newcommand{\goesto}{$\rightarrow$}
\newcommand{\DA}[1]{{\bf #1}}
\newcommand{\twocol}[1]{\multicolumn{2}{c|}{#1}}

\begin{table}[H]
\begin{singlespace}
\caption{Changes in DA Labeling Associated with Listening.
  Changes are denoted as original label (transcript-only)\goesto new label (transcript
  + listening).  In Round I, labeler was unaware of DAs of interest; in
  Round II, labeler was biased toward the most frequent change from
  Round I (Backchannel\goesto Agreement). Labels are from original DA
  classes (as listed in Appendix A): \DA{b}=Backchannel,
  \DA{aa}=Agreement, \DA{sv}=Statement-opinion,
  \DA{sd}=Statement-non-opinion.}

\label{tab:relabeling}

\begin{tabular}{|l|cc|cc|}
\hline
                & \twocol{Round I}              & \twocol{Round II}     \\
\hline
\hline
Goal of study   & \twocol{\parbox{2in}{\centering Which DAs change most?}} &
 \twocol{\parbox[t]{2in}{\centering What is upper bound for DA-specific change rate?}} \\
Task focus      & \twocol{All DAs}              & \twocol{\DA{b} and \DA{aa}} \\
Relabeling time & \twocol{20 total hrs}         & \twocol{10 hrs} \\
Number of conversations & \twocol{44}                 & \twocol{19 (not in Round I)} \\
\hline
Changed DAs (\%) & 114/5857     & 1.95\%        & 114/4148      & 2.75\% \\
\hline
Top changes (\% of total changes) & & & & \\
\quad \DA{b}\goesto\DA{aa}
                & 43/114        & 37.7\%        & 72/114        & 63.2\% \\
\quad \DA{sv}\goesto\DA{sd}
                & 22/114        & 19.3\%        & 2/114         & 1.75\% \\
\quad \DA{sd}\goesto\DA{sv}
                & 17/114        & 14.9\%        & 0             & 0\% \\
\quad Other changes &           & $<$3\% each   &       & $<$8\% each   \\
\hline
Change rate, relative to total DAs & & & & \\
\quad \DA{b}\goesto\DA{aa}
                 & 43/5857      & 0.73\%        & 72/4148       & 1.74\% \\
\quad Other changes & 71/5857   & 1.21\%        & 42/4148       & 1.01\% \\
\hline
Change rate, relative to DA priors
                 & & & & \\
\quad \DA{b}\goesto\DA{aa} / \DA{b}
                & 43/986        & 4.36\%        & 72/690        & 10.43\% \\
\quad Non-\DA{b}/\DA{aa}\goesto Non-\DA{b}/\DA{aa} / Non-\DA{b}/\DA{aa}
                & 57/4544       & 1.25\%        & 11/3180       & 0.35\% \\
\hline
\end{tabular}
\end{singlespace}
\end{table}

Only 114 changes were made in Round I, for an overall rate
of change of under 2\%. Given that attention was divided over all
DAs in this round, the most meaningful information from Round I is not
the overall rate of changes, which is expected to be conservative, but
rather the distribution of types of changes.  The most prominent
change made after listening was the conversion of Backchannels
(\DA{b}) to Agreements (\DA{aa}).
Details on the prosodic cues associated with this change are
described elsewhere \cite{JurafskyEtAl:acl}.
As the table shows for top
changes, this change accounted for 43, or 37.7\%, of the 114 changes made;
the next most frequent change (within the two different original
Statement labels) accounted for less than 20\% of the changes.%
\footnote{In addition, many of the \DA{sd}\goesto\DA{sv} changes were in
  fact an indirect result of \DA{b}\goesto\DA{aa} changes for the
  following utterance.}  The salience of the \DA{b}\goesto\DA{aa}
changes is further seen after normalizing the number of changes by the
DA priors. On this measure, \DA{b}\goesto\DA{aa} changes occur for over
4\% of original \DA{b} labels. In contrast, the normalized rates for the
second and third most frequent types of changes in Round I were 22/989
(2.22\%) for \DA{sv}\goesto\DA{sd} and 17/2147 (0.79\%) for
\DA{sd}\goesto\DA{sv}.  For all changes not involving either \DA{b} or
\DA{aa}, the rate was only about 1\%.  A complete list of recall and
precision rates by DA type (where labels after listening are used as
reference labels, and labels from transcripts alone are used as
hypothesized labels), can be found in Appendix~B.

To address the issue of attention to changing the original labels, we
ran a second round of relabeling (Round II). Since
\DA{b}\goesto\DA{aa} changes were clearly the most salient from Round I,
we discussed these changes with the labeler, and then asked her to
relabel additional conversations with attention to these changes. Thus,
we expected her to focus relatively more attention on
\DA{b}\goesto\DA{aa} in Round II (although she was instructed also to
label any other glaring changes). We viewed Round II as a way to obtain
an upper bound on the DA-specific change rate, since
\DA{b}\goesto\DA{aa} changes were the most frequently occurring changes
after listening, and since the labeler was biased toward focusing
attention on these changes. For Round II, we used a completely separate
set of data from Round I, to avoid
confounding the relabeling procedure. 
The overall distribution of DAs was similar to that in the set used in Round I.

As shown in Table~\ref{tab:relabeling}, the number of changes made in
Round II was the same (by coincidence) as in Round I.  However, since
there were fewer total utterances in Round I, the rate of change relative to total
DAs increased from Round I to Round II.  In Round II,
\DA{b}\goesto\DA{aa} changes greatly increased from Round I, both relative to total DAs and relative to DA-specific priors. At the same time, other types of
changes decreased from Round I to Round II.

The most important result from Round II is the rate of
\DA{b}\goesto\DA{aa} changes relative to the prior for the \DA{b}
class. This value was about 10\%, and is a reasonable estimate of the
upper bound on DA changes for any particular class from listening,
since it is unlikely that listening would
affect other DAs more than it did Backchannels, given both the predominance of
\DA{b}\goesto\DA{aa} changes in Round I, and the fact that the labeler
was biased to attend to \DA{b}\goesto\DA{aa} changes in Round
II. These results suggest that at least 90\% of the utterances
in any of our originally labeled DA classes are likely to be marked
with the same DA label after listening, and that for most other DAs
this value should be considerably higher.  Therefore, although our
transcript-only labels contained some errors, based on the results of
the relabeling experiments we felt that it was reasonable to use the
transcript-only labels as estimates of after-listening labels.

\subsubsection{Interlabeler reliability}
Interlabeler reliability on our main (transcript-only) set of
annotations was assessed using the Kappa statistic
\cite{Cohen:60,SieCas:88,Carletta96}, or the ratio of the proportion of times
that raters agree (corrected for chance agreement) to the maximum
proportion of times that the rates could agree (corrected for chance
agreement). Kappa computed for the rating of the original 42 classes
was 0.81, which is considered high for this type of task.  {\em Post
hoc} grouping of the ratings using the seven main classes just
described yielded a Kappa of 0.85.

\subsection{Training and Test Sets}

We partitioned the available data into three subsets for training and
testing.  The three subsets were not only disjoint but also shared
no speakers.  The {\em training set} (TRN) contained 1794 conversation
sides; its acoustic waveforms were used to train decision trees, while
the corresponding transcripts served as training data for the
statistical language models used in word-based DA classification.  The {\em
  held-out set} (HLD) contained 436 conversation sides; it was used to
test tree performance as well as DA classification based on true words.  A much
smaller {\em development test set} (DEV) consisting of 38 matched
conversation sides (19 conversations) was used to perform experiments
involving automatic word recognition, as well as 
corresponding experiments based on prosody and true words.%
\footnote{The DEV set was so called because of its role in the WS97
  projects that focused on word recognition.}
The TRN and HLD sets
contained single, unmatched conversation sides, but since no discourse
context was required for the studies reported here this was not a
problem.  The three corpus subsets with their statistics are summarized
in Table~\ref{tab:corpora}.

\begin{table}[H]
\begin{singlespace}
\caption{Summary of Corpus Training and Test Subsets}
\label{tab:corpora}
\begin{center}
\begin{tabular}{|l|l|r|r|r|}
\hline
Name    & Description & Sides & Utterances & Words \\
\hline
\hline
TRN     & Training set          & 1794 & 166K & 1.2M \\
\hline
HLD     & Held-out test set     & 436 & 32K & 231K \\
\hline
DEV     & Development test set  & 19  & 4K & 29K \\
\hline
\end{tabular}
\end{center}
\end{singlespace}
\end{table}

\subsection{Dialog Act Segmentation}

In a fully automated system, DA classification presupposes the ability
to also find the boundaries between utterances.
In spite of extensive work on this problem in recent years,
to our knowledge there are currently no systems that reliably perform utterance
segmentation for spontaneous conversational speech
when the true words are not known.
For this work we did not want to confound the issue of DA classification
with DA segmentation; thus, we used utterance boundaries marked by
human labelers according to the LDC annotation guidelines
described in \namecite{Meteer95}. To keep results using different knowledge sources
comparable, these DA boundaries were also made explicit for purposes of speech
recognition and language modeling.\footnote{%
Note that the very notion of utterances and utterance boundaries
is a matter of debate and subject to research \cite{TraumHeeman:96}.
We adopted a pragmatic approach by choosing a pre-existing segmentation
for this rather large corpus.}

The utterance boundaries were marked between words. To estimate the
locations of the boundaries in the speech waveforms, a forced alignment
of the acoustic training data was merged with the training
transcriptions containing the utterance boundary annotations marked by
the LDC. This yielded word and pause times of the training data with
respect to the acoustic segmentations.  By using these word times along
with the linguistic segmentation marks, the start and end times for
linguistic segments were found.

This technique was not perfect, however.  One problem is
that many of the words included in the linguistic transcription had been
excised from the acoustic training data.  Some speech segments were
considered not useful for acoustic training and thus had been excluded
deliberately.  In addition, the alignment program was allowed to skip
words at the beginning and end of an acoustic segment if there was
insufficient acoustic evidence for the word. This caused misalignments in the
context of highly reduced pronunciations or for low-energy speech, both
of which are frequent in Switchboard.  
Errors in the boundary times for DAs crucially affect the prosodic
analyses, since prosodic features are extracted assuming that the boundaries
are reasonably correct. Incorrect estimates affect the accuracy of
global features (e.g., DA duration) and may render local features
meaningless (e.g., F0 measured at the supposed end of the utterance).
Since features for DAs with known problematic end estimates would be
misleading in the prosodic analyses, they were omitted from all of our
TRN and HLD data. The time boundaries of the
DEV test set, however, were carefully handmarked for other purposes, so
we were able to use exact values for this test set. Overall, we were
missing 30\% of the utterances in the TRN and HLD sets because of
problems with time boundaries; this figure was higher for particular
utterance types, especially for short utterances such as backchannels,
for which as much as 45\% of the utterances were affected. 
Thus, the DEV set was mismatched with respect to the TRN and HLD sets 
in terms of the percentage of utterances affected by problematic segmentations.

\subsection{Prosodic Features}

The prosodic database included a variety of features that could be
computed automatically without reference to word information. In
particular, we attempted to have good coverage of features and feature
extraction regions that were expected to play a role in the three focused
analyses mentioned in the Introduction: detection of Questions,
Agreements, and Incomplete Utterances. Based on the literature on
question intonation \cite{Vaissiere:83,Haan1EtAl:97,Haan2EtAl:97}, we
expected Questions to show rising F0 at the end of the utterance,
particularly for Declarative and Yes-No Questions. Thus, F0 should be a
helpful cue for distinguishing Questions from other long DAs such as
Statements.  Many Incomplete Utterances give the impression of being cut
off prematurely, so the prosodic behavior at the end of such an
utterance may be similar to that of the middle of a normal utterance.
Specifically, energy can be expected to be higher at the end of an
abandoned utterance compared to energy at the end of a completed one.
In addition, unlike most completed utterances, the F0 contour at the end
of an Incomplete Utterance is neither rising nor falling. 
We expected Backchannels to differ from Agreements by the amount of
effort used in speaking. Backchannels function to acknowledge another
speaker's contributions without taking the floor, whereas Agreements
assert an opinion. We therefore expected Agreements to have higher
energy, greater F0 movement, and a higher likelihood of accents and
boundary tones than Backchannels.

\subsubsection{Duration features}
Duration was expected to be a good cue for discriminating Statements and
Questions from DAs functioning to manage the dialog (e.g.,
Backchannels), although this difference is also encoded to some extent
in the language model.  In addition to the duration of the utterance in
seconds, we included features correlated with utterance duration, but
based on frame counts conditioned on the value of other feature types,
as shown in Table~\ref{tab:feats-dur}.

\begin{table}[H]
\begin{singlespace}
\caption{Duration Features}
\label{tab:feats-dur}
\begin{center}
\begin{tabular}{|l|l|}
\hline
Feature Name & Description \\
\hline
\hline
 Duration & \\
   \quad ling\_dur & \PB{  duration of utterance}\\
\hline
 Duration-pause & \\
   \quad ling\_dur\_minus\_min10pause & \PB{  ling\_dur minus sum of duration of all pauses of at least 100 ms}\\
   \quad cont\_speech\_frames & \PB{  number of frames in continuous speech regions ($>1$ s, ignoring pauses $< 10$ frames)}\\
\hline
 Duration-correlated F0-based counts & \\
   \quad f0\_num\_utt & \PB{  number of frames with F0 values in utterance (prob\_voicing=1)}\\
   \quad f0\_num\_good\_utt & \PB{  number of F0 values above f0\_min (f0\_min = .75*f0\_mode)}\\
   \quad regr\_dur & \PB{  duration of F0 regression line (from start to end point, includes voiceless frames)}\\
   \quad regr\_num\_frames & \PB{  number of points used in fitting F0 regression line (excludes voiceless frames)}\\
   \quad numacc\_utt & \PB{  number of accents in utterance from event recognizer}\\
   \quad numbound\_utt & \PB{  number of boundaries in utterance from event recognizer}\\
\hline
\end{tabular}
\end{center}
\end{singlespace}
\end{table}

The duration-pause set of features computes duration, ignoring pause
regions. Such features may be useful if pauses are unrelated to
DA classification. (If pauses are relevant, however, this should be
captured by the pause features described in the next section.)  The
F0-based count features reflect either the number of frames or
recognized intonational events (accents or boundaries) based on F0
information (see F0 features, below). The first four of these features
capture time in speaking by using knowledge about the presence and location
of voiced frames, which may be more robust for our data than relying on
pause locations from the alignments.  The last two features are intended
to capture the amount of information in the utterance, by counting
accents and phrase boundaries.  Duration-normalized versions of many of
these features are included under their respective feature type in the
following sections.

\subsubsection{Pause features}
To address the possibility that hesitation could provide a cue to the
type of DA, we included features intended to reflect the degree of
pausing, as shown in Table~\ref{tab:feats-pau}.  To obtain pause
locations we used information available from forced alignments; however,
this was only for convenience (the alignment information was included in our
database for other purposes). In principle, pause locations can be
detected by current recognizers with high accuracy without knowledge of
the words. Pauses with durations below 100 milliseconds (10 frames) were
excluded since they are more likely to reflect segmental information
than hesitation. Features were normalized to remove the inherent
correlation with utterance duration. The last feature 
provides a more global measure of pause behavior, including pauses 
during which the other speaker was talking. 
The measure counts only those speech frames occurring in regions of at
least 1 second of continuous speaking.  The window was run over the
conversation (by channel), writing out a binary value for each frame; the
feature was then computed based on the frames within a particular DA.

\begin{table}[H]
\begin{singlespace}
\caption{Pause Features}
\label{tab:feats-pau}
\begin{center}
\begin{tabular}{|l|l|}
\hline
Feature Name & Description \\
\hline
\hline
 min10pause\_count\_n\_ldur & \PB{  number of pauses of at least 10 frames (100 ms) in utterance, normalized by duration of utterance}\\
 total\_min10pause\_dur\_n\_ldur & \PB{  sum of duration of all pauses of at least 10 frames in utterance, normalized by duration of utterance}\\
 mean\_min10pause\_dur\_utt & \PB{  mean pause duration for pauses of at least 10 frames in utterance}\\
 mean\_min10pause\_dur\_ncv & \PB{  mean pause duration for pauses of at least 10 frames in utterance, normalized by same in convside}\\
 cont\_speech\_frames\_n & \PB{  number of frames in continuous speech regions ($>1$ s, ignoring pauses $< 10$ frames) normalized by duration of utterance}\\
\hline
\end{tabular}
\end{center}
\end{singlespace}
\end{table}

\subsubsection{F0 features}
F0 features, shown in
Table~\ref{tab:feats-f0}, included both raw values (obtained from
ESPS/Waves+) and values from a linear regression (least-squares fit)
to the frame-level F0 values.

\begin{table}[H]
\begin{singlespace}
\caption{F0 Features}
\label{tab:feats-f0}
\begin{center}
\begin{tabular}{|l|l|}
\hline
Feature Name & Description \\
\hline
\hline
 f0\_mean\_good\_utt & \PB{  mean of F0 values included in f0\_num\_good\_utt}\\
 f0\_mean\_n & \PB{  difference between mean F0 of utterance and mean F0 of convside for F0 values $>$ f0\_min}\\
 f0\_mean\_ratio & \PB{  ratio of F0 mean in utterance to F0 mean in convside}\\
 f0\_mean\_zcv & \PB{  mean of good F0 values in utterance normalized by mean and st dev of good F0 values in convside}\\
 f0\_sd\_good\_utt & \PB{  st dev of F0 values included in f0\_num\_good\_utt}\\
 f0\_sd\_n & \PB{  log ratio of st dev of F0 values in utterance and in convside}\\
 f0\_max\_n & \PB{  log ratio of max F0 values in utterance and in convside}\\
 f0\_max\_utt & \PB{  maximum F0 value in utterance (no smoothing)}\\
 max\_f0\_smooth & \PB{  maximum F0 in utterance after median smoothing of F0 contour}\\
 f0\_min\_utt & \PB{  minimum F0 value in utterance (no smoothing); can be below f0\_min}\\
\hline
 f0\_percent\_good\_utt & \PB{  ratio of number of good F0 values to number of F0 values in utterance}\\
\hline
 utt\_grad & \PB{  least-squares all-points regression over utterance}\\
 pen\_grad & \PB{  least-squares all-points regression over penultimate region}\\
 end\_grad & \PB{  least-squares all-points regression over end region}\\
 end\_f0\_mean & \PB{  mean F0 in end region}\\
 pen\_f0\_mean & \PB{  mean F0 in penultimate region}\\
 abs\_f0\_diff & \PB{  difference between mean F0 of end and penultimate regions}\\
 rel\_f0\_diff & \PB{  ratio of F0 of end and penultimate regions}\\
 norm\_end\_f0\_mean & \PB{  mean F0 in end region normalized by mean and st dev of F0 from convside}\\
 norm\_pen\_f0\_mean & \PB{  mean F0 in penultimate region normalized by mean and st dev from convside}\\
 norm\_f0\_diff & \PB{  difference between mean F0 of end and penultimate regions, normalized by mean and st dev of F0 from convside}\\
 regr\_start\_f0 & \PB{  first F0 value of contour, determined by regression line analysis}\\
\hline
 finalb\_amp & \PB{  amplitude of final boundary (if present), from event recognizer}\\
 finalb\_label & \PB{  label of final boundary (if present), from event recognizer}\\
 finalb\_tilt & \PB{  tilt of final boundary (if present), from event recognizer}\\
 numacc\_n\_ldur & \PB{  number of accents in utterance from event recognizer, normalized by duration of utterance}\\
 numacc\_n\_rdur & \PB{  number of accents in utterance from event recognizer, normalized by duration of F0 regression line}\\
 numbound\_n\_ldur & \PB{  number of boundaries in utterance from event recognizer, normalized by duration of utterance}\\
 numbound\_n\_rdur & \PB{  number of boundaries in utterance from event recognizer, normalized by duration of F0 regression line}\\
\hline
\end{tabular}
\end{center}
\end{singlespace}
\end{table}

To capture overall pitch range, mean F0 values were calculated over all
voiced frames in an utterance. To normalize differences in F0 range over
speakers, particularly across genders, utterance-level values were
normalized with respect to the mean and standard deviation of F0 values
measured over the whole conversation side.  F0 difference values were
normalized on a log scale. The standard deviation in F0 over an
utterance was computed as a possible measure of expressiveness over the
utterance. Minimum and maximum F0 values, calculated after median
smoothing to eliminate spurious values, were also included for this
purpose.%
\footnote{A more linguistically motivated measure of the maximum F0
  would be to take the F0 value at the RMS maximum of the sonorant
  portion of the nuclear-accented syllable in the phrase (e.g.,
  \opencite{HirschbergNakatani:96}).  However, our less sophisticated
  measure of pitch range was used as an approximation because we did not
  have information about the location of accents or phrase boundaries
  available.}

We included parallel measures that used only ``good'' F0 values, or
values above a threshold (f0\_min) estimated as the bottom of a
speaker's natural F0 range.  The f0\_min can be calculated in two ways.
For both methods, a smoothed histogram of all the calculated F0 values
for a conversation side is used to find the F0 mode.  The true f0\_min
comes from the minimum F0 value to the left of this mode.  Because the
histogram can be flat or not sufficiently smoothed, the algorithm could be
fooled into choosing a value greater than the true minimum.  A simpler
way to estimate the f0\_min takes advantage of the fact that values below the
minimum typically result from pitch halving. Thus, a good estimate of
f0\_min is to take the point at 0.75 times the F0 value at the mode of
the histogram.  This measure closely approximates the true f0\_min, and
is more robust for use with the Switchboard data.%
\footnote {We thank David Talkin for suggesting this method.}  The
percentage of ``good'' F0 values was also included to measure
(inversely) the degree of creaky voice or vocal fry.

The rising/falling behavior of pitch contours is a good cue to their
utterance type. We investigated several ways to measure this
behavior. To measure overall slope, we calculated the gradient of a
least-squares fit regression line for the F0 contour. While this gives
an adequate measure for the overall gradient of the utterance, it is not
always a good indicator of the type of rising/falling behavior in which we are
most interested. Rises at the end can be swamped by the declination
of the preceding part of the contour, and hence the overall
gradient for a contour can be falling. We therefore marked two special
regions at the end of the contour, corresponding to the last 200 milliseconds
(end region) and the 200 milliseconds previous to that (penultimate
region). For each of these regions we measured the mean F0 and gradient,
and used the differences between these as features. The starting value
in the regression line was also included as a potential cue to F0
register (the actual first value is prone to F0 measurement error).

In addition to these F0 features, we included intonational-event
features, or features intended to capture local pitch accents and
phrase boundaries. The event features were obtained using the event
recognizer described in \namecite{TaylorEtAl:97}. The event detector
uses an HMM approach to provide an intonational segmentation of an
utterance, which gives the locations of pitch accents and boundary
tones. When compared to human intonation transcriptions of
Switchboard,%
\footnote{As labeled by the team of students at Edinburgh; see
  Acknowledgments.} this system correctly identifies 64.9\% of events,
but has a high false alarm rate, resulting in an accuracy of 31.7\%.

\subsubsection{Energy features}
We included two types of energy features, as shown in
Table~\ref{tab:feats-nrg}. The first set of features 
was computed based on standard RMS energy.
Because our data were recorded from telephone handsets with various noise
sources (background noise as well as channel noise), we also included a
signal-to-noise ratio (SNR) feature to try to capture the energy from the
speaker. SNR values were calculated using the SRI recognizer with a
Switchboard-adapted front end \cite{NeuWei:94,NeuWei:95}. 
Values were calculated over the entire conversation side, and those
extracted from regions of speech were used to find a cumulative
distribution function (CDF) for the conversation.  The frame-level SNR
values were then represented by their CDF value to normalize the
SNR values across speakers and conversations.

\begin{table}[H]
\begin{singlespace}
\caption{Energy Features}
\label{tab:feats-nrg}
\begin{center}
\begin{tabular}{|l|l|}
\hline
Feature Name & Description \\
\hline
 utt\_nrg\_mean & \PB{  mean RMS energy in utterance}\\
 abs\_nrg\_diff & \PB{  difference between mean RMS energy of end and penultimate regions}\\
 end\_nrg\_mean & \PB{  mean RMS energy in end region}\\
 norm\_nrg\_diff & \PB{  normalized difference between mean RMS energy of end and penultimate regions }\\
 rel\_nrg\_diff & \PB{  ratio of mean RMS energy of end and penultimate regions}\\
\hline
 snr\_mean\_utt & \PB{  mean SNR (CDF value) in utterance}\\
 snr\_sd\_utt & \PB{  st dev of SNR values (CDF values) in utterance}\\
 snr\_diff\_utt & \PB{  difference between maximum SNR and minimum SNR in utterance}\\
 snr\_min\_utt & \PB{  st dev of SNR values (CDF values) in utterance}\\
 snr\_max\_utt & \PB{  maximum SNR value (CDF values) in utterance}\\
\hline
\end{tabular}
\end{center}
\end{singlespace}
\end{table}

\subsubsection{Speaking rate (enrate) features}
We were also interested in overall speaking rate.  However, we needed a
measure that could be run directly on the signal, since our features
could not rely on word information.  For this purpose, we experimented
with a signal processing measure, ``enrate''
\cite{MorganEtAl:97},
which estimates a syllable-like rate by looking at the energy in the speech signal after preprocessing. Studies comparing enrate values to values based on hand-transcribed syllable rates for Switchboard show a correlation of about .46 for the version of the software used in the present work.%
\footnote{We thank Nelson Morgan, Eric Fosler-Lussier, and Nikki Mirghafori for
  allowing us to use the software and note that the measure has since
  been improved (mrate), with correlations increasing to about .67
  as described in \namecite{MorganFosler:98}.}

The measure can be run over the entire signal, but
because it uses a large window, values are less meaningful if
significant pause time is included in the window. 
We calculated frame-level values over a 2-second speech interval.  The
enrate value was calculated for a 25-millisecond frame window with a window step size of 200 milliseconds.  Output values were calculated every 10 milliseconds to
correspond to other measurements.  We included pauses of less than 1
second and ignored speech regions of less than 1 second, where pause
locations were determined as described earlier. 

If the end of a speech segment was approaching, meaning that the
2-second window could not be filled, no values were written out.  The
enrate values corresponding to particular utterances were then extracted
from the conversation-side values.  This way, if utterances were
adjacent, information from surrounding speech regions could be used to
get enrate values for the beginnings and ends of utterances that
normally would not fill the 2-second speech window. Features computed
for use in tree-building are listed in Table~\ref{tab:feats-rat}.

\begin{table}[H]
\begin{singlespace}
\caption{Speaking Rate Features}
\label{tab:feats-rat}
\begin{center}
\begin{tabular}{|l|l|}
\hline
Feature Name & Description \\
\hline
mean\_enr\_utt & \PB{  mean of enrate values in utterance}\\
mean\_enr\_utt\_norm & \PB{  mean\_enr\_utt normalized by mean enrate in conversation side}\\
stdev\_enr\_utt & \PB{  st dev of enrate values in utterance}\\
min\_enr\_utt & \PB{  minimum enrate value in utterance}\\
max\_enr\_utt & \PB{  maximum enrate value in utterance}\\
\hline
\end{tabular}
\end{center}
\end{singlespace}
\end{table}

\subsubsection{Gender features}
As a way to check the effectiveness of our F0
normalizations we included the gender of the speaker. It is also
possible that features could be used differently by men and women, 
even after appropriate normalization for pitch range differences.
We also included the gender of the listener to check for a possible
sociolinguistic interaction between the conversational dyad and the ways in which speakers employ
different prosodic features.

\subsection{Decision Tree Classifiers}

For our prosodic classifiers, we used CART-style decision trees
\cite{Breiman83}. Decision trees can be trained to perform classification using a combination of discrete and
continuous features, and can be inspected to gain an understanding of
the role of different features and feature combinations.

We downsampled our data (in both training and testing) to obtain an
equal number of datapoints in each class. Although an inherent
drawback is a loss of  power in the analyses due to fewer datapoints,
downsampling was warranted for two reasons.  First, as noted
earlier, the distribution of frequencies for our DA classes was severely
skewed.  Because decision trees split according to an entropy criterion,
large differences in class size wash out any effect of the features
themselves, causing the tree not to split.  By downsampling to equal
class priors we assure maximum sensitivity to the features. A second
motivation for downsampling was that by training our classifiers on a
uniform distribution of DAs, we facilitated integration with other
knowledge sources (see section on Integration).
After expanding the tree with questions, the training
algorithm used a tenfold cross-validation procedure to avoid overfitting the
training data. Leaf nodes were successively pruned if they failed to
reduce the entropy in the cross-validation procedure.

We report tree performance using two metrics, {\em accuracy} and
{\em efficiency}. Accuracy is the number of correct classifications
divided by the total number of samples.  Accuracy is based on hard
decisions; the classification is that class with the highest posterior
probability.  Because we downsampled to equal class priors, the chance
performance for any tree with N classes is 100/N\%.  For any particular
accuracy level, there is a trade-off between recall and false alarms. In
the real world there may well be different costs to a false positive
versus a false negative in detecting a particular utterance type. In the
absence of any model of how such costs would be assigned for our data,
we report results assuming equal costs to these errors.
 
Efficiency measures the relative reduction in entropy between the prior
class distribution and the posterior distribution predicted by the tree.
Two trees may have the same classification
accuracy, but the tree that more closely approximates the probability
distributions of the data (even if there is no effect on decisions) has
higher efficiency (lower entropy).  Although accuracy and efficiency are
typically correlated, the relationship between the measures is not
strictly monotonic since efficiency looks at probability distributions
and accuracy looks only at decisions.

\subsection{Dialog Act Classification from Word Sequences}

Two methods were used for classification of DAs from word information.
For experiments using the correct words $W$, we needed to compute the
likelihoods $P(W | U)$ for each DA or utterance type $U$, i.e., the
probability with which $U$ generates the word sequence $W$. The predicted
DA type would then be the one with maximum likelihood.
To estimate these probabilities,
we grouped the transcripts of the training corpus
by DA type, and trained a standard trigram language model using backoff
smoothing \cite{Katz:87} for each DA.  This was done for the
original 42 DA categories, yielding 42 DA-specific language models.
Next, for experiments involving a DA class $C$ comprising several of the
original DAs $U_1$, $U_2$, \ldots, $U_n$, we combined the DA likelihoods
in a weighted manner:
\[
        P(W | C) = P(W | U_1) P(U_1 | C) + \ldots + P(W | U_n) P(U_n | C)
\]
Here, $P(U_1 | C)$, \ldots, $ P(U_n | C)$ are the relative frequencies
of the various DAs within class $C$.

For experiments involving (necessarily imperfect) automatic word
recognition, we were given only the acoustic information $A$.  We
therefore needed to compute acoustic likelihoods $P(A | U)$, i.e., the
probability that utterance type $U$ generates the acoustic
manifestation $A$.  In principle, this can be accomplished by considering
all possible word sequences $W$ that might have generated the acoustics
$A$, and summing over them:
\[
        P(A | U) = \sum_{W} P(A | W) P(W | U)
\]
Here $P(W | U)$ is estimated by the same DA-specific language models
as before, and $P(A | W)$ is the acoustic score of a speech recognizer,
expressing how well the acoustic observations match the word sequence $W$.
In practice, however, we could only consider a finite number of potential
word hypotheses $W$; in our experiments we generated the 2500 most likely 
word sequences for each utterance, and carried out the above summation 
over only those sequences.
The recognizer used was a state-of-the-art HTK large-vocabulary 
recognizer, which nevertheless had a word error rate of 41\% on the
test corpus.\footnote{%
Note that the summation over multiple word hypotheses
is preferable to the more straightforward approach of looking 
at only the one best hypothesis and treating it as the actual words
for the purpose of DA classification.}

\subsection{Integration of Knowledge Sources}

To use multiple knowledge sources for DA classification, i.e., prosodic
information as well as other, word-based evidence, we combined tree
probabilities $P(U | F)$ and word-based likelihoods $P(W | U)$
multiplicatively.  This approach can be justified as follows.  The
likelihood-based classifier approach dictates choosing the DA with the
highest likelihood based on both the prosodic features $F$ and the words
$W$, $P(F, W | U)$.  To make the computation tractable, we assumed,
similar to \namecite{TaylorEtAl:LS98}, that
the prosodic features are independent of the words once conditioned on
the DA. We recognize, however, that this assumption is a simplification.%
\footnote{Utterance length is one feature for which this independence
assumption is clearly violated.
Utterance length is represented by a prosodic feature (utterance duration)
as well as implicitly in the DA-specific language models.
\namecite{FinkeEtAl:aaai98} suggest a way to deal with this particular
problem by conditioning the language models on utterance length.}
Our prosodic model averages over all examples of a particular DA; it is
``blind'' to any differences in prosodic features that correlate with
word information.  For example, statements about a favorite sports team
use different words than statements about personal finance, and the two
different types of statements tend to differ prosodically (e.g.,
in animation level as reflected by overall pitch range). 
In future work,
such differences could potentially be captured by using more sophisticated
models designed to represent semantic or topic information.  For practical
reasons, however, we consider our prosodic models independent of the
words once conditioned on the DA, i.e.:

\begin{eqnarray*}
        P(F, W | U) & = & P(W | U) P(F | W , U) \\
                & \approx & P(W | U) P(F | U) \\
                & \propto & P(W | U) P(U | F)
\end{eqnarray*}

The last line is justified because, as noted earlier, we trained the
prosodic trees on downsampled data or a uniform distribution of DA
classes. According to Bayes' Law, the required likelihood $P(F | U)$ equals
$P(U | F) P(F) / P(U)$.  The second factor, $P(F)$, is the same for all
DA types $U$, and $P(U)$ is equalized by the downsampling procedure.
Hence, the probability estimated by
the tree, $P(U | F)$, is proportional to the likelihood $P(F | U)$.
Overall, this justifies multiplying $P(W | U)$ and $P(U | F)$.%
\footnote{In practice
  we needed to adjust the dynamic ranges of the two probability
  estimates by finding a suitable exponential weight $\lambda$, to make

\[
        P(F, W | U)  \propto P(W | U) P(F | U)^\lambda\quad. 
\]
}

\section{Results and Discussion}

We first examine results of the prosodic model for a seven-way
classification involving all DAs. We then look at results from a
words-only analysis, to discover potential subtasks for which prosody
could be particularly helpful. The words-only analysis reveals that even
if correct words are available, certain DAs tend to be misclassified.
We examine
the potential role of prosody for three such subtasks: (1) the detection
of Questions, (2) the detection of Agreements, and (3) the detection of
Incomplete Utterances.  In all analyses we seek to understand the
relative importance of different features and feature types, as well as
to determine whether integrating prosodic information with a language
model can improve classification performance overall.

\subsection{Seven-Way Classification}

We applied the prosodic model first to a seven-way classification task
for the full set of DAs: Statements, Questions, Incomplete Utterances,
Backchannels, Agreements, Appreciations, and Other.  Note that ``Other''
is a catch-all class representing numerous heterogeneous DAs
that occurred infrequently in our data. Therefore we do not expect this
class to have consistent features.
As described in the Method section, data were downsampled to equal class
sizes to avoid confounding results with information from prior frequencies of
each class.  Because there are seven classes, chance accuracy
for this task is 100/7\% or 14.3\%.  For simplicity, we assumed equal
cost to all decision errors, i.e., to all possible confusions among the
seven classes.

A tree built using the full database of features described earlier
yields a classification accuracy of 41.15\%.  This gain in accuracy is
highly significant by a binomial test, $p<.0001$.  If we are interested in
probability distributions rather than decisions, we can look at the
efficiency of the tree, or the relative reduction in entropy over the
prior distribution.  By using the all-features prosodic tree for this
seven-way classification, we reduce the number of bits necessary to
describe the class of each datapoint by 16.8\%.

The all-features tree is large (52 leaves), making it difficult to
interpret the tree directly.  In such cases we found it useful to
summarize the overall contribution of different features by using a measure of 
``feature usage'', which is proportional to the number of times a feature
was queried in classifying the datapoints. The measure thus accounts for
the position of the feature in the tree: features used higher in the
tree have greater usage values than those lower in the tree since there are
more datapoints at the higher nodes. The measure is normalized to sum to
1.0 for each tree.  Table~\ref{tab:featuse-clust-152} lists usage by
feature type.

\begin{table}[H]
\begin{singlespace}
\caption{Feature Usage for Main Feature Types in Seven-Way Classification}
\label{tab:featuse-clust-152}
\begin{center}
\begin{tabular}{|l|l|}
\hline
Feature & Usage \\
Type    &   \\
\hline
\hline
Duration & 0.554 \\
\hline
F0 & 0.126 \\
\hline
Pause & 0.121 \\
\hline
Energy & 0.104 \\
\hline
Enrate & 0.094 \\
\hline
\end{tabular}
\end{center}
\end{singlespace}
\end{table}

Table~\ref{tab:featuse-clust-152} indicates that when all features are
available, a duration-related feature is used in more than half of the
queries. Notably, gender features are not used at all; this supports
the earlier hypothesis that, as long as features are appropriately
normalized, it is reasonable to create gender-independent prosodic
models for these data.  A summary of individual feature usage, as shown in
Table~\ref{tab:featuse-raw-152}, reveals that the raw duration feature
(ling\_dur)---which is a ``free'' measure in our work since we assumed
locations of utterance boundaries---accounted for only 14\% of the
queries in the tree; the remaining queries of the 55\% accounted for by
duration features were those associated with the computation of F0- and
pause-related information.  Thus, the power of duration for the seven-way
classification comes largely from measures involving computation of
other prosodic features.  The most-queried feature, regr\_num\_frames
(the number of frames used in computing the F0 regression line), may be
better than other duration measures at capturing actual speech portions
(as opposed to silence or nonspeech sounds), and may be better than
other F0-constrained duration measures (e.g., f0\_num\_good\_utt) because of
a more robust smoothing algorithm.  We can also note that the high
overall rate of F0 features given in Table~\ref{tab:featuse-raw-152}
represents a summation over many different individual features.

\begin{table}[H]
\begin{singlespace}
\caption{Feature Usage for Seven-Way (all DAs) Classification}
\label{tab:featuse-raw-152}
\begin{center}
\begin{tabular}{|l|l|l|}
\hline
Feature & Feature & Usage \\
Type    &         &   \\
\hline
\hline
Duration & regr\_num\_frames & 0.180 \\
\hline
Duration & ling\_dur & 0.141 \\
\hline
Pause & cont\_speech\_frames\_utt\_n & 0.121 \\
\hline
Enrate & stdev\_enr\_utt & 0.081 \\
\hline
Enrate & ling\_dur\_minus\_min10pause & 0.077 \\
\hline
Pause & cont\_speech\_frames\_utt & 0.073 \\
\hline
Energy & snr\_max\_utt & 0.049 \\
\hline
Energy & snr\_mean\_utt & 0.043 \\
\hline
Duration & regr\_dur & 0.041 \\
\hline
F0 & f0\_mean\_zcv & 0.036 \\
\hline
F0 & f0\_mean\_n & 0.027 \\
\hline
Duration & f0\_num\_good\_utt & 0.021 \\
\hline
Duration & f0\_num\_utt & 0.019 \\
\hline
F0 & norm\_end\_f0\_mean & 0.017 \\
\hline
F0 & numacc\_n\_rdur & 0.016 \\
\hline
F0 & f0\_sd\_good\_utt & 0.015 \\
\hline
Energy & mean\_enr\_utt & 0.009 \\
\hline
F0 & f0\_max\_n & 0.006 \\
\hline
Energy & snr\_sd\_utt & 0.006 \\
\hline
Energy & rel\_nrg\_diff & 0.005 \\
\hline
Enrate & mean\_enr\_utt\_norm & 0.004 \\
\hline
F0 & regr\_start\_f0 & 0.003 \\
\hline
F0 & finalb\_amp & 0.003 \\
\hline
\end{tabular}
\end{center}
\end{singlespace}
\end{table}

\begin{figure}[H]
\begin{singlespace}
\begin{center}
\includebargraph{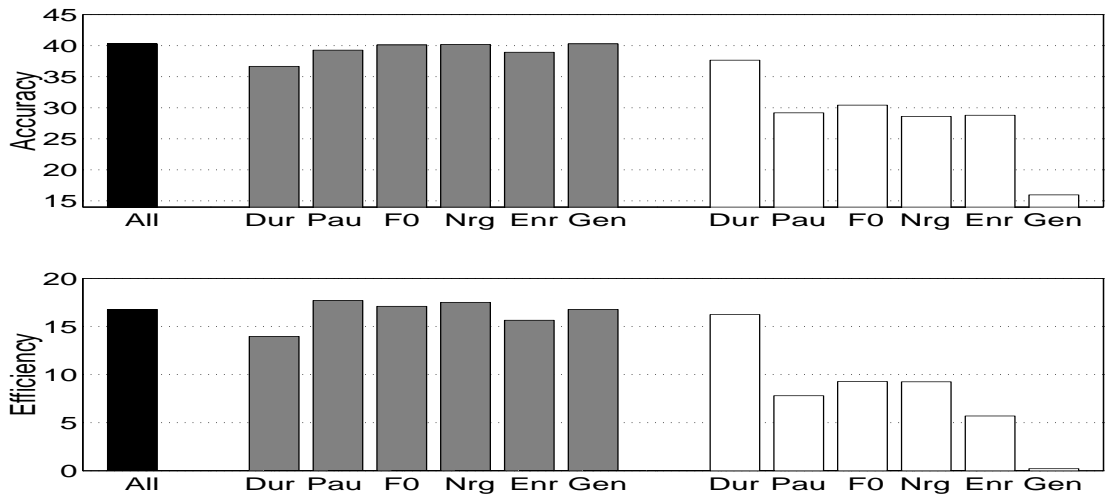} 
\end{center}
\caption{Performance of prosodic trees using different feature sets for the 
  classification of all seven DAs (Statements, Questions, Incomplete
  Utterances, Backchannels, Agreements, Appreciations, Other).  $N$ (number of samples in each class)=391. Chance accuracy=14.3\%. Gray bars=exclude feature
  type; white bars=include only feature type.  Dur=Duration, Pau=Pause,
  F0=Fundamental frequency, Nrg=Energy, Enr=Enrate (speaking rate),
  Gen=Gender features.}
\label{fig:bar152-all7}
\end{singlespace}
\end{figure}

Since we were also interested in feature importance,
individual trees were built using the leave-one-out method, in which
the feature list is systematically modified and a new tree is built for
each subset of allowable features.
It was not feasible to leave out individual features because of the
large set of features used; we therefore left out groups of features
corresponding to the feature types as defined in the Method section.  We also
included a matched set of ``leave-one-in'' trees for each of the feature
types (i.e., trees for which all {\em other} feature types were removed)
and a single leave-two-in tree, built {\em post hoc}, which made
available the two feature types with highest accuracy from the
leave-one-in analyses.  Note that the defined feature lists specify the
features {\em available} for use in building a particular prosodic
model; whether or not features are {\em actually} used is determined by
the tree learning algorithm and depends on the data.
Figure~\ref{fig:bar152-all7} shows results for the set of
leave-one-out and leave-one-in trees, with the all-features tree
provided for comparison.  The upper graph indicates accuracy
values; the lower graph shows efficiency values.  Each bar indicates a
separate tree.

We first tested whether there was any significant loss in leaving out a
feature type,  by doing pairwise comparisons between the all-features tree and
each of the leave-one-out trees.%
\footnote{To test whether one tree (A) was significantly better than
  another (B), we counted the number of test instances on which A and B
  differed, and on how many instances A was correct but B was not; we
  then applied a Sign test to these counts.}  Although trees with more
features to choose from typically perform better than those with fewer
features, additional features can hurt performance.  The greedy tree-growing
algorithm does not look ahead to determine the overall best feature set,
but rather seeks to maximize entropy reduction locally at each split.
This limitation of decision trees is another motivation for conducting
the leave-one-out analyses.  Since we cannot predict the direction of
change for different feature sets, comparisons on tree results were
conducted using two-tailed tests.

Results showed that the only two feature types whose removal caused a
significant reduction in accuracy were duration ($p<0.0001$) and
enrate ($p<0.05$). The enrate-only tree, however, yields accuracies on
par with other feature types whose removal did not affect overall
performance; this suggests that the contribution of enrate in the
overall tree may be through interactions with other features. All of
the leave-one-in trees were significantly less accurate than the
all-features tree. Although the tree using only duration achieved an
accuracy close to that of the all-features tree, it was still
significantly less accurate by a Sign test ($p<0.01$).  Adding F0
features (the next-best feature set in the leave-one-in trees) did not
significantly improve accuracy over the duration-only tree alone,
suggesting that for this task the two feature types are highly
correlated.  Nevertheless, for each of the leave-one-in trees, all feature
types except gender yielded accuracies significantly above chance by a
binomial test ($p<.0001$ for the first five trees).  The gender-only
tree was slightly better than chance by either a one- or a two-tailed
test.%
\footnote{It is not clear here whether a one- or two-tailed
test is more appropriate.  Trees typically should not do worse than chance;
however,
  because they minimize entropy and not accuracy, the accuracy can fall
  slightly below chance.} However, this was most likely due to a
difference in gender representation across classes.

Taken together, these results suggest that there is
considerable redundancy in the features for DA classification, since
removing one feature type at a time (other than duration) makes little
difference to accuracy.  Results also suggest, however, that features are
not perfectly correlated; there must be considerable interaction among
features in classifying DAs, because trees using only individual feature
types are significantly less accurate than the all-features tree.

Finally, duration is clearly of primary importance to this
classification. This is not surprising, as the task involves a seven-way
classification including longer utterances (such as Statements) and very
brief ones (such as Backchannels like ``uh-huh'').
Two questions of further interest regarding duration, however, are (1)
will a prosody model that uses mostly duration add anything to a
language model (in which duration is implicitly encoded), and (2) is
duration useful for other tasks involving classification of DAs similar
in length? Both questions are addressed in the following analyses.

As just discussed, the all-features tree (as well as  others
including only subsets of feature types) provides significant
information for the seven-way classification task.  Thus, if one were
to use only prosodic information (no words or context), this is the
level of performance resulting for the case of equal class
frequencies. To explore whether the prosodic information could be of
use when lexical information is also available, we integrated the tree
probabilities with likelihoods from our DA-specific trigram language models
built from the same data. 
For simplicity, integration results are reported only for the
all-features tree in this and all further analyses, although as noted
earlier this is not guaranteed to be the optimal tree.

Since our trees were trained with uniform class priors, we combined tree
probabilities $P(U | F)$ with the word-based likelihoods $P(W | U)$
multiplicatively, as described in the Integration section.%
\footnote{The relative weight assigned to the prosodic and the word
  likelihoods was optimized on the test set due to lack of an additional
  held-out data set.  Although in principle this could bias results, we
  found empirically that similar performance is obtained using a range
  of weighting values; this is not surprising since only a single
  parameter is estimated.}

The integration was performed
separately for each of our two test sets (HLD and DEV), and within the
DEV set for both transcribed and recognized words. Results are shown in
Table~\ref{tab:integration-152}.  Classification performance is shown for
each of the individual classifiers, as well as for the optimized
combined classifier.

\begin{table}[H]
\begin{singlespace}
\caption{Accuracy of Individual and Combined Models for Seven-Way Classification}
\label{tab:integration-152}
\begin{center}
\begin{tabular}{|l|c|c|c|}
\hline
Knowledge & HLD Set & DEV Set & DEV Set \\
Source    & true words  &  true words & N-best output \\         
\hline
\hline
samples           &  2737 &  287   & 287    \\
chance (\%)       & 14.29 & 14.29  & 14.29 \\ 
tree   (\%)       & 41.15 & 38.03  & 38.03 \\
words   (\%)      & 67.61 & 70.30  & 58.77 \\
words+tree (\%)   & 69.98 & 71.14  & 60.12 \\
\hline
\end{tabular}
\end{center}
\end{singlespace}
\end{table}

As shown, for all three analyses, adding information from the tree to
the word-based model improved classification accuracy. Although
the gain appears modest in absolute terms,
for the HLD test set it was highly significant by a Sign test,%
\footnote{One-tailed, because model integration assures no loss in
  accuracy.} $p<.001$.  For the smaller DEV test set, the improvements
did not reach significance; however, the pattern of results suggests that
this is likely to be due to the small sample
size. It is also the case that the tree model does not perform as well
for the DEV as the HLD set. This is not attributable to small sample
size, but rather to a mismatch between the DEV set and the training data
involving how data were segmented, as noted in the Method section. The
mismatch in particular affects duration features, which were important
in these analyses as discussed earlier.  Nevertheless, word-model results
are lower for N-best than for true words in the DEV data, while by
definition the tree results stay the same.  We see that accordingly,
integration provides a larger win for the recognized than for the true
words. Thus, we would expect that results for recognized words for the HLD set
(if they could be obtained) should show an even larger win than the win
observed for the true words in that set.

These results provide an answer to one of the questions posed earlier:
does prosody provide an advantage over words if the
prosody model uses mainly duration? The results indicate that the answer
is yes.  Although the number of words in an utterance is highly
correlated with duration, and word counts are represented implicitly by
the probability of the end-of-utterance marker in a language model, a
duration-based tree model still provides added benefit over words alone.
One reason may be that duration (reflected by the various features we
included) is simply a better predictor of DA than is word count.
Another independent possibility is that the benefit from the tree
model comes from its ability to threshold feature values directly and
iteratively.

\subsection{Dialog Act Confusions Based on Word Information}

\begin{figure}[H]
\begin{singlespace}
\begin{center}
  \includeLMbargraph{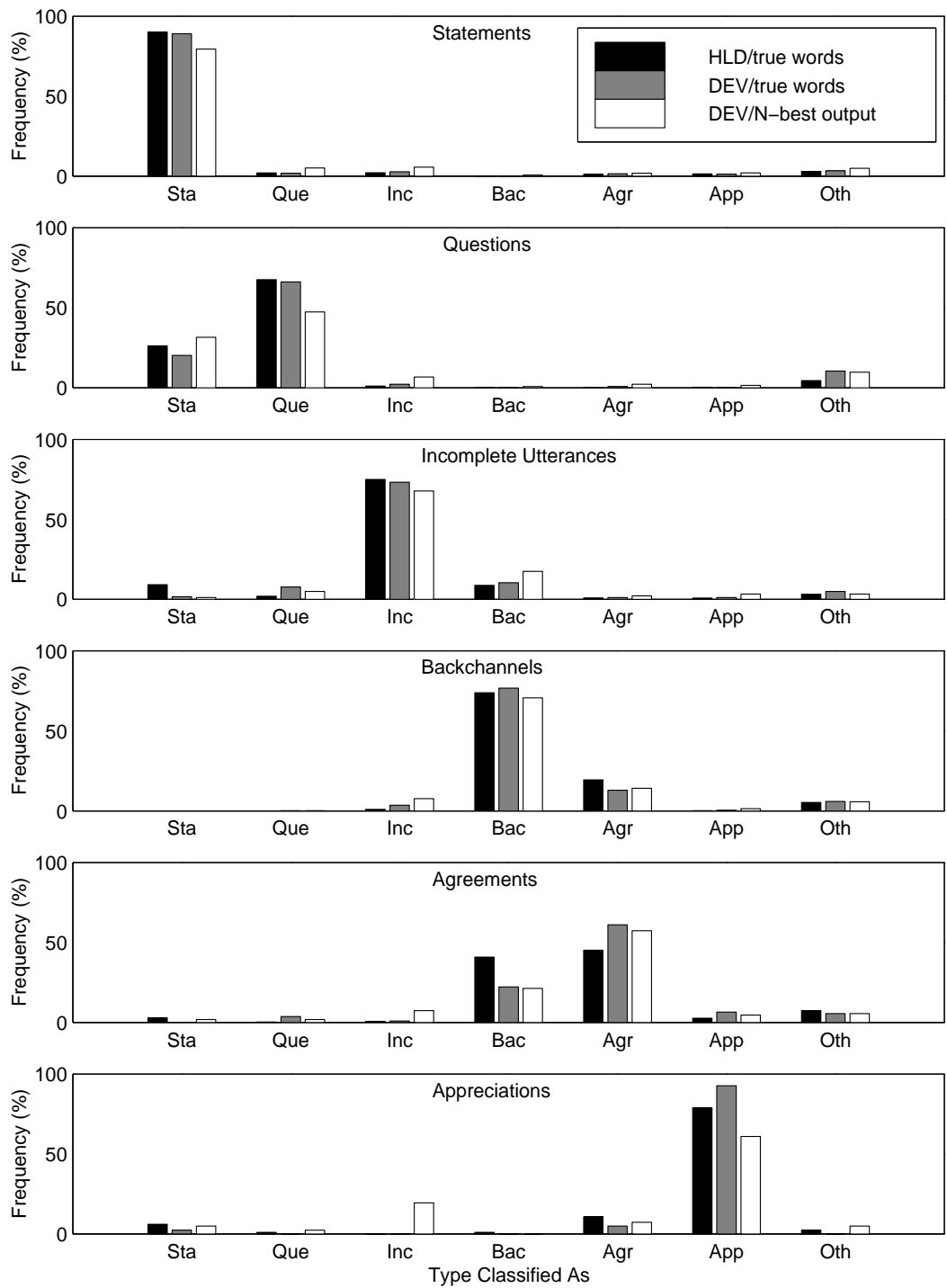}
\end{center}
\caption{Classification of DAs based on word trigrams only}
\label{fig:lmconf}
\end{singlespace}
\end{figure}

Next we explored additional tasks for which prosody could aid DA
classification. Since our trees allow N-ary classification, the logical
search space of possible tasks was too large to explore systematically.
We therefore looked to the language model to guide us in identifying
particular tasks of interest.  Specifically, we were interested in DAs
that tended to be misclassified even given knowledge of the true words.
We examined the pattern of confusions made when our seven DAs
were classified using the language model alone.  Results are shown in
Figure~\ref{fig:lmconf}.
Each subplot represents the data for one actual DA.%
\footnote{Because of the heterogeneous makeup of the ``Other'' DA class,
    we were not {\em per se} interested in its pattern of confusions,
    and hence the graph for that data is not shown.}  Bars reflect the
    normalized rate at which the actual DA was classified as each of
    the seven possible DAs, in each of the three test conditions
    (HLD, DEV/true, and DEV/N-best).

As shown, classification is excellent for the Statement class, with few
misclassifications even when only the recognized words are used.%
\footnote{The high classification rate for Statements by word
  information was a prime motivation for downsampling our data in order
  to examine the inherent contribution of prosody, since as noted in the
  Method section, Statements make up most of the data in this corpus.}
For the remaining DAs, however, misclassifications occur at considerable
rates.%
\footnote{Exact classification accuracy values for the various DAs shown
  in Figure~\ref{fig:lmconf} are provided in the text as needed for the
  subtasks examined, i.e. under ``words'' in
  Tables~\ref{tab:integration-150}, \ref{tab:integration-143}, and
  \ref{tab:integration-145}.}  Classification of Questions is a case in
point: even with true words, Questions are often misclassified as
Statements (but not vice versa), and this pattern is exaggerated when
testing on recognized as opposed to true words.  The asymmetry is
partially attributable to the presence of declarative Questions.  The
effect associated with recognized words appears to reflect a high rate
of missed initial ``do'' in our recognition output, as discovered in
independent error analyses \cite{JuretalTR:98}.  For both Statements and
Questions, however, there is little misclassification involving the
remaining classes. This probably reflects the length distinction as well
as the fact that most of the propositional content in our corpus
occurred in Statements and Questions, whereas other DAs generally served
to manage the communication---a distinction likely to be reflected in
the words.  Thus, our first subtask was to examine the role of
prosody in the classification of Statements and Questions.  A second
problem visible in Figure~\ref{fig:lmconf} is the detection of
Incomplete Utterances. Even with true words, classification of these
DAs is at only 75.0\% accuracy. Knowing whether or not a DA is complete
would be particularly useful for both language modeling and
understanding. Since the misclassifications are distributed over the set
of DAs, and since logically any DA can have an incomplete counterpart,
our second subtask was to classify a DA as either incomplete or
not-incomplete (all other DAs).  A third notable pattern of confusions
involves Backchannels and explicit Agreements. This was an expected
difficult discrimination as discussed earlier, since the two classes
share words such as ``yeah'' and ``right''.  In this case, the
confusions are considerable in both directions.

\subsection{Subtask 1: Detection of Questions}

\begin{figure}[H]
\begin{singlespace}
\begin{center}
\includetree{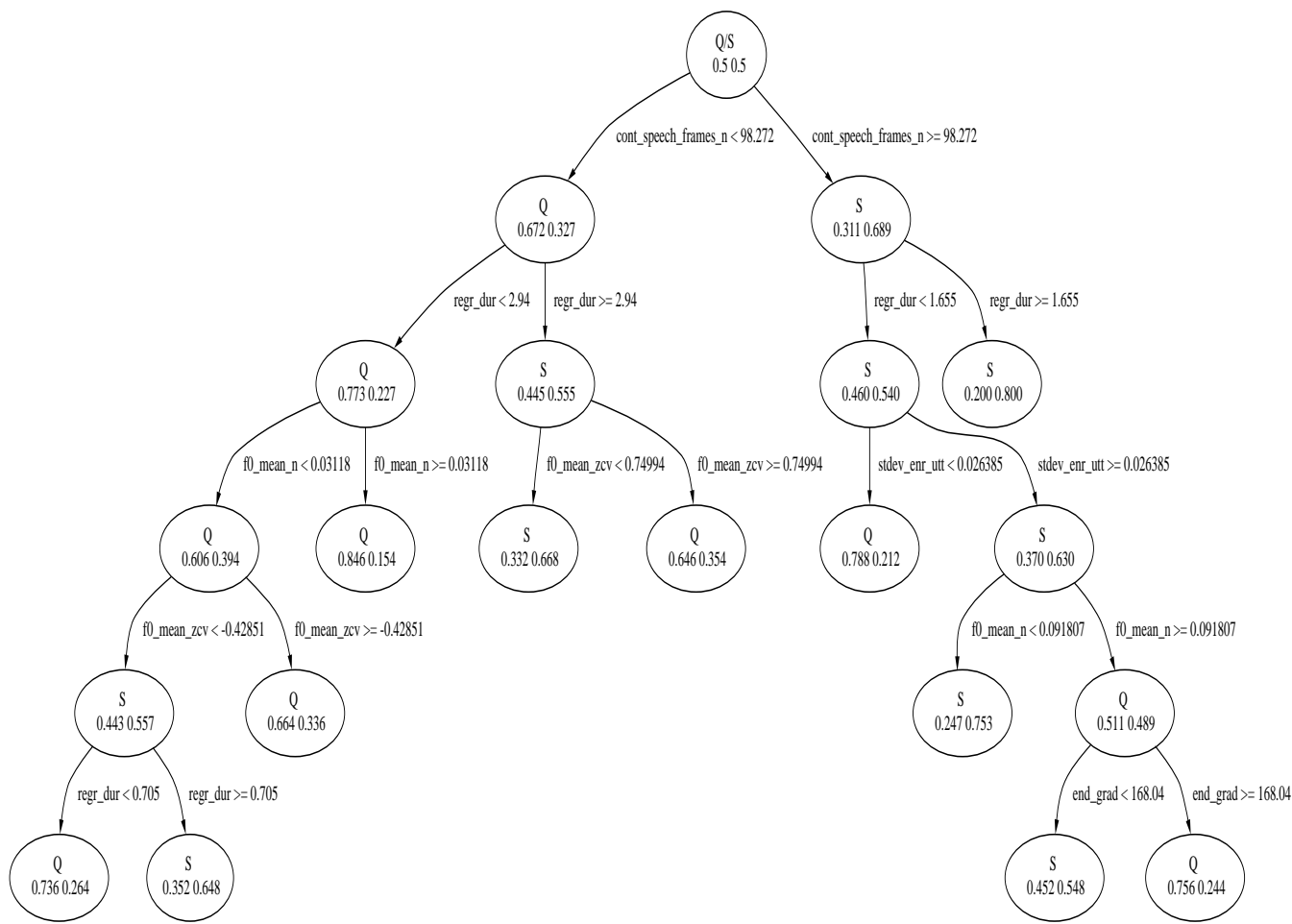}
\end{center}
\caption{Decision tree for the classification of Statements (S) and Questions (Q)}
\label{fig:tree-150}
\end{singlespace}
\end{figure}

As illustrated in the previous section, Questions in our corpus
were often misclassified as Statements based on words alone. Based on
the literature, we hypothesized that prosodic features, particularly
those capturing the final F0 rise typical of some Question types in
English, could play a role in reducing the rate of misclassifications.
To investigate the hypothesis, we built a series of classifiers using
only Question and Statement data.  We first examined results for an
all-features tree, shown in Figure~\ref{fig:tree-150}.  Each node
displays the name of the majority class, as well as the posterior
probability of the classes (in the class order indicated in the top
node). Branches are labeled with the name of the feature and threshold
value determining the split. The tree yields an accuracy of 74.21\%,
which is significantly above the chance level of 50\% by a binomial
test, $p<.0001$; the tree reduces the number of bits necessary to describe the
class of each datapoint by 20.9\%.

\subsubsection{Feature importance}
The feature usage of the tree is summarized in
Table~\ref{tab:featuse-raw-150}.  As predicted, F0 features help differentiate
Questions from Statements, and in the expected direction (Questions have
higher F0 means and higher end gradients than Statements).  What was
not obvious at the outset is the extent to which other features also cue this
distinction. In the all-features tree, F0 features comprise only about
28\% of the total queries. Two other features, regr\_dur and
cont\_speech\_frames, are each queried more often than the F0 features
together. Questions are shorter in duration (from starting to ending
voiced frame) than Statements.  They also have a lower percentage of
frames in continuous speech regions than Statements. Further inspection
suggests that the pause feature in this case (and also most
likely for the seven-way classification discussed earlier) indirectly
captures information about turn boundaries surrounding the DA of interest.
Since our speakers were recorded on different channels, the end of one
speaker's turn is often associated with the onset of a long pause
(during which the other speaker is talking). Furthermore, long pauses
reduce the frame count for the continuous-speech-frames feature enrate
measure because of the windowing described earlier. Therefore, this measure
reflects the timing of continuous speech spurts across speakers, and is
thus different in nature from the other pause features that look only
inside an utterance.

\begin{table}[H]
\begin{singlespace}
\caption{Feature Usage for Classification of Questions and Statements}
\label{tab:featuse-raw-150}
\begin{center}
\begin{tabular}{|l|l|l|}
\hline
Feature & Feature & Usage \\
Type    &         &   \\
\hline
\hline
Duration & regr\_dur & 0.332 \\
\hline
Pause & cont\_speech\_frames\_n  & 0.323 \\
\hline
F0 &  f0\_mean\_n  &  0.168 \\
\hline
F0 &  f0\_mean\_zcv &  0.088 \\
\hline
Enrate & stdev\_enr\_utt &  0.065 \\
\hline
F0 &  end\_grad &  0.024 \\
\hline
\end{tabular}
\end{center}
\end{singlespace}
\end{table}

\begin{figure}[H]
\begin{singlespace}
\begin{center}
\includebargraph{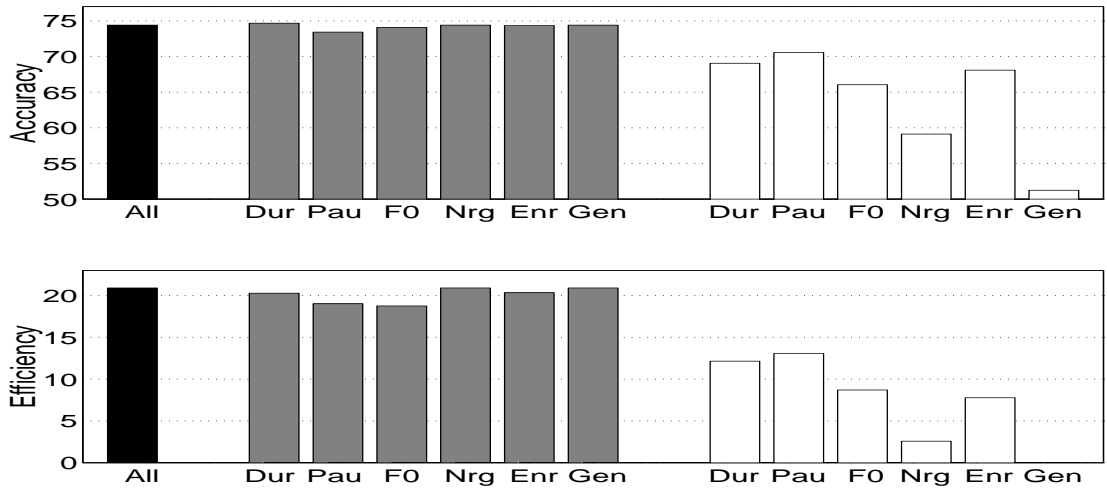} 
\end{center}
\caption{Performance of prosodic trees using different feature sets for the 
  classification of Statements and Questions. $N$ for each class=926.
  Chance accuracy = 50\%. Gray bars=exclude feature type; white
  bars=include only feature type.  Dur=Duration, Pau=Pause,
  F0=Fundamental frequency, Nrg=Energy, Enr=Enrate (speaking rate),
  Gen=Gender features.}
\label{fig:bar150-sq}
\end{singlespace}
\end{figure}

To further examine the role of features, we built additional trees using
partial feature sets. Results are summarized in
Figure~\ref{fig:bar150-sq}. As suggested by the leave-one-out trees,
there is no significant effect on accuracy when any one of the feature
types is removed.  Although we predicted that Questions should differ
from Statements mainly by intonation, results indicate that a tree with
no F0 features achieves the same accuracy as a tree with all features
for the present task.  Removal of all pause features, which resulted in
the largest drop in accuracy, yields a tree with an accuracy of 73.43\%,
which is not significantly different from that of the all-features tree
($p=.2111$, n.s.). Thus, if any feature type is removed, other feature
types compensate to provide roughly the same overall accuracy.  However,
it is not the case that the main features used are perfectly correlated,
with one substituting for another that has been removed.  Inspection of
the leave-one-out tree reveals that upon removal of a feature type, new
features (features, and feature types, that never appeared in the
all-features tree) are used. Thus, there is a high degree of
redundancy in the features that differentiate Questions and Statements,
but the relationship among these features and the allowable feature
sets for tree building is complex.

\begin{figure}[H]
\begin{singlespace}
\begin{center}
\strut\psfig{file=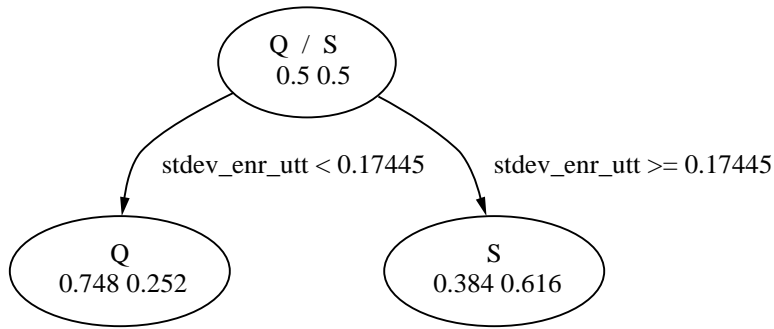,angle=270,width=4in,height=1.7in}
\end{center}
\caption{Decision tree for the classification of Statements (S) and Questions (Q), using
  only enrate features}
\label{fig:tree150-1r-sq}
\end{singlespace}
\end{figure}

Inspection of the leave-one-in tree results in
Figure~\ref{fig:bar150-sq} indicates, not surprisingly, that the feature
types most useful in the all-features analyses (duration and pause)
yield the highest accuracies for the leave-one-in analyses (all of which
are significantly above chance, $p<.0001$). It is interesting, however,
that enrate, which was used only minimally in the all-features tree,
allows classification at 68.09\%,
which is 
better than that of the F0-only tree.
Furthermore, the enrate-only classifier is a mere shrub: as shown in
Figure~\ref{fig:tree150-1r-sq}, it splits only once, on an
{\em unnormalized} feature that expresses simply the variability in enrate
over the utterance.  As noted in the Method section, enrate is expected
to correlate with speaking rate, although for this work we were not able
to investigate the nature of this relationship.  However, the result has
interesting potential implications.  Theoretically, it suggests that
absolute speaking rate may be less important for DA classification than
variation in speaking rate over an utterance; a theory of
conversation should be able to account for the lower variability in
questions than in statements.  For applications, results suggest that
the inexpensive enrate measure could be used alone to help distinguish
these two types of DAs in a system in which other feature types are not
available.

We ran one further analysis on question classification. The aim was to
determine the extent to which our grouping of different kinds of
questions into one class affected the features used in question
classification. As described in the Method section, our Question class
included Yes-No Questions, Wh-questions, and Declarative Questions.
These different types of questions are expected to differ in their
intonational characteristics \cite{Quirk85,Weber:93,Haan1EtAl:97,Haan2EtAl:97}.
Yes-No Questions and Declarative Questions typically involve a final F0
rise; this is particularly true for Declarative Questions whose function is not conveyed
syntactically. Wh-Questions, on the other hand, often fall in F0, as do
Statements.

\begin{figure}[H]
\begin{singlespace}
\begin{center}
\includetree{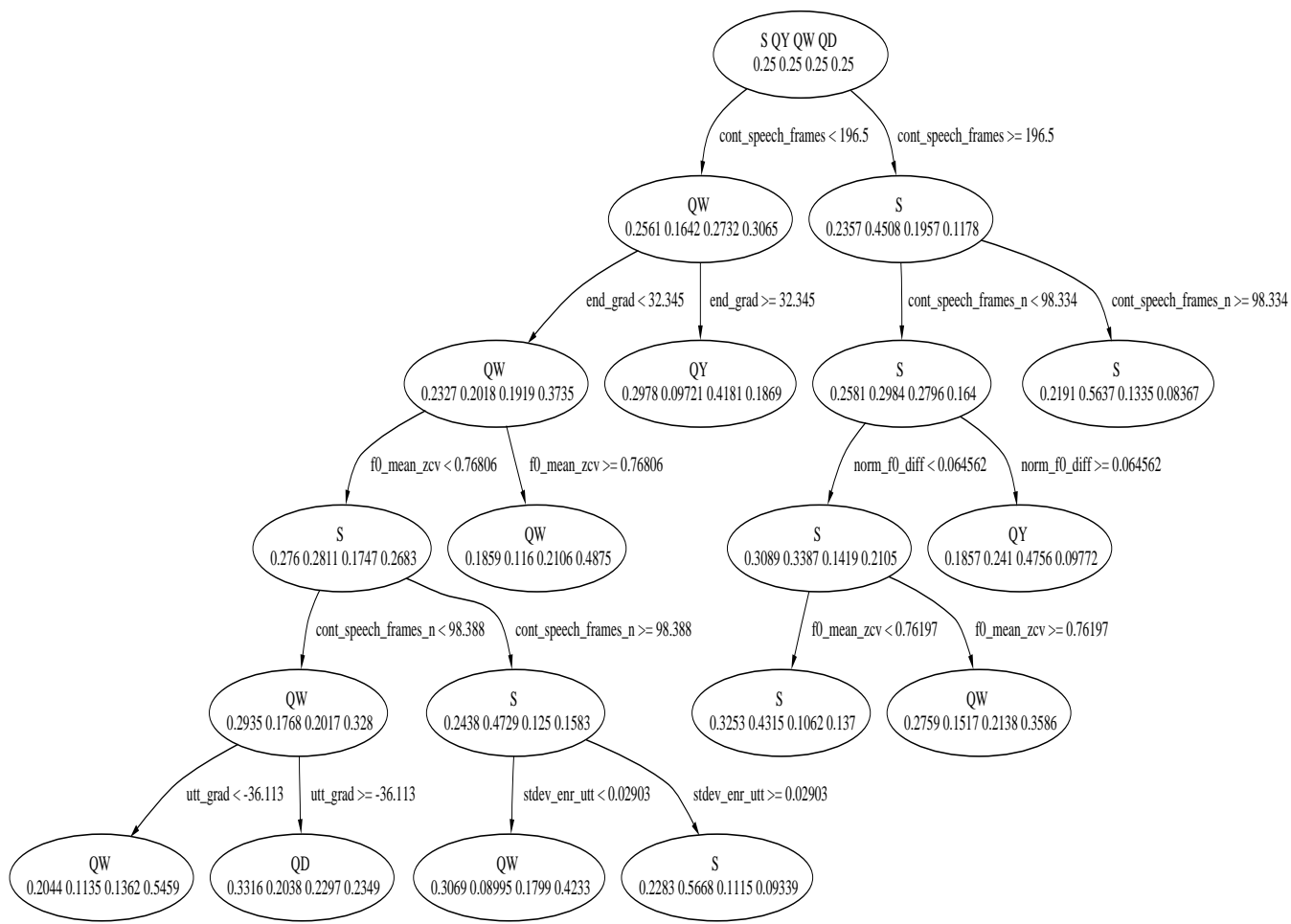}
\end{center}
\caption{Decision tree for the classification of Statements (S), Yes-No Questions (QY), Wh-Questions (QW), and Declarative Questions (QD)}
\label{fig:tree-158}
\end{singlespace}
\end{figure}

We broke down our Question class into the originally coded Yes-No
Questions, Wh-Questions, and Declarative Questions, and ran a four-way
classification along with Statements. The resulting all-features tree
is shown in Figure~\ref{fig:tree-158}, and a summary of the feature
usage is provided in Table~\ref{tab:featuse-clust-158}.

\begin{table}[H]
\begin{singlespace}
\caption{Feature Usage for Main Feature Types in Classification of Yes-No Questions, Wh-Questions, Declarative Questions, and Statements}
\label{tab:featuse-clust-158}
\begin{center}
\begin{tabular}{|l|l|}
\hline
Feature  & Usage \\
Type     &  \\
\hline
\hline
F0 & 0.432 \\
\hline
Duration & 0.318 \\
\hline
Pause & 0.213 \\
\hline
Enrate & 0.037 \\
\hline
\end{tabular}
\end{center}
\end{singlespace}
\end{table}

The tree achieves an accuracy of 47.15\%, a highly significant increase
over chance accuracy (25\%) by a binomial test, $p<.0001$.  Unlike the
case for the grouped Question class, the most queried feature type is
now F0. Inspection of the tree reveals that the pattern of results is
consistent with the literature on question intonation. Final rises
(end\_grad, norm\_f0\_diff, and utt\_grad) are associated with Yes-No
and Declarative Questions, but not with Wh-Questions. Wh-Questions show
a higher average F0 (f0\_mean\_zcv) than Statements.

\begin{figure}[H]
\begin{singlespace}
\begin{center}
\includebargraph{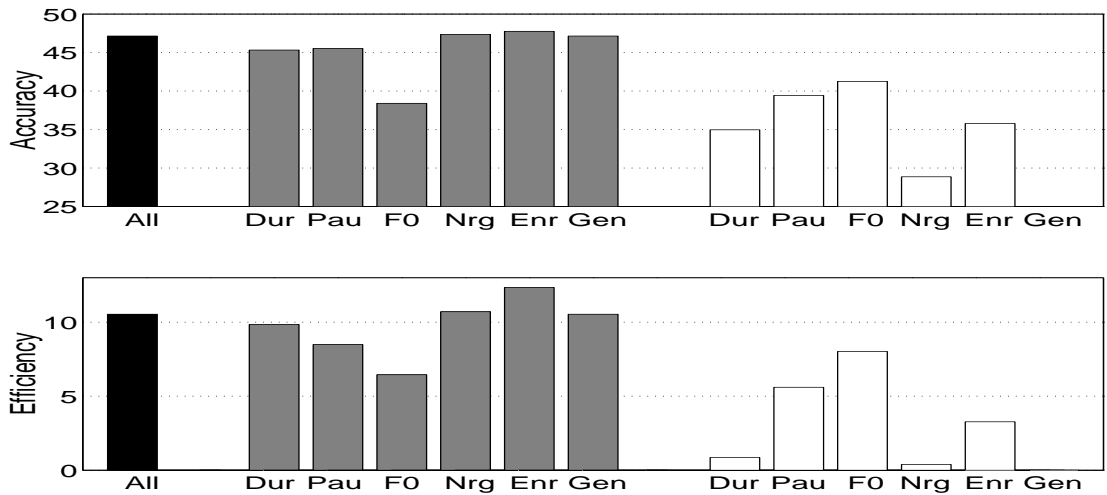}
\end{center}
\caption{Performance of prosodic trees using different feature sets
  for the classification of Statements, Yes-No Questions,
  Wh-Questions, and Declarative Questions. $N$ for each
  class=123. Chance=25\%. Gray bars=exclude feature type; white
  bars=include only feature type.  Dur=Duration, Pau=Pause,
  F0=Fundamental frequency, Nrg=Energy, Enr=Speaking rate, Gen=Gender
  features.  }
\label{fig:bar158-S-Qy-Qw-Qd}
\end{singlespace}
\end{figure}

To further assess feature importance, we again built trees after
selectively removing feature types. Results are shown in
Figure~\ref{fig:bar158-S-Qy-Qw-Qd}. 

In contrast to Figure~\ref{fig:bar150-sq}, in which accuracy was
unchanged by removal of any single feature type,
the data in Figure~\ref{fig:bar158-S-Qy-Qw-Qd} show a
sharp reduction in accuracy when F0 features are removed. This result is
highly significant by a Sign test ($p<.001$, two-tailed) despite the
small amount of data in the analyses, resulting from downsampling to
the size of the least frequent question subclass.
For all other feature types, there was
no significant reduction in accuracy when the feature type was removed.
Thus, F0 plays an important role in question detection, but because
different kinds of questions are signaled in different ways
intonationally, combining questions into a single class as in the
earlier analysis smoothes over some of the distinctions.  In particular,
the grouping tends to conceal the features associated with the final F0
rise (probably because the rise is averaged in with final falls).

\subsubsection{Integration with language model}

To answer the question of whether prosody can aid Question
classification when word information is also available, tree
probabilities were combined with likelihoods from our DA-specific
trigram language models, using an optimal weighting factor.  Results
were computed for the two test sets (HLD and DEV) and within the DEV
set for both transcribed and recognized words. The outcome is shown in
Table~\ref{tab:integration-150}.

\begin{table}[H]
\begin{singlespace}
\caption{Accuracy of Individual and Combined Models for the Detection of Questions}
\label{tab:integration-150}
\begin{center}
\begin{tabular}{|r|c|c|c|}
\hline
Knowledge & HLD Set & DEV Set & DEV Set \\
Source    & true words  &  true words & N-best output \\         
\hline
\hline
 samples           &  1852 & 266    &  266      \\
 chance (\%)       & 50.00 & 50.00  & 50.00     \\ 
 tree   (\%)       & 74.21 & 75.97  & 75.97     \\
 words   (\%)      & 83.65  & 85.85 & 75.43      \\
 words+tree (\%)   & 85.64  & 87.58 & 79.76     \\  
\hline
\end{tabular}
\end{center}
\end{singlespace}
\end{table}

The prosodic tree model yielded accuracies significantly better than
chance for both test sets ($p<.0001$). The tree alone was also
more accurate than the recognized words alone for this task.
Integration yielded consistent improvement over the words alone. The
larger HLD set showed a highly significant gain in accuracy for the
combined model over the words-only model, $p<.001$ by a Sign test.
Significance tests were not meaningful for the DEV set because of a lack
of power given the small sample size; however, the pattern of
results for the two sets is similar (the spread is
greatest for the recognized words) and therefore suggestive.

\subsection{Subtask 2: Detection of Incomplete Utterances}

A second problem area in the words-only analyses was the classification 
of Incomplete Utterances.   Utterances labeled as incomplete in our
work included three different main phenomena:%
\footnote{In addition, the class included a variety of utterance types
  deemed ``uninterpretable'' because of premature cut-off.}

\begin{singlespace}
\begin{center}
\begin{tabular}{|lccl|}
\hline
  Turn exits:       &               & (A) & We have young children. \\
                    & $\rightarrow$ & (A) & So \ldots \\
                    &               & (B) & Yeah, that's tough then. \\
\hline
Other-interruptions: & $\rightarrow$& (A) & We eventually --- \\
                     &              & (B) & Well you've got to start somewhere. \\
\hline
Self-interruptions: & $\rightarrow$ & (A) & And they were definitely --- \\
(repairs)           &               & (A) & At halftime they were up by four.\\
\hline
\end{tabular}
\end{center}
\end{singlespace}
     
Although the three cases represent different phenomena,
they are similar in that in each case the utterance could
have been completed (and coded as the relevant type) but was not.
An all-features tree built for the classification of Incomplete Utterances
and all other classes combined (Non-Incomplete) yielded an accuracy 
of 72.16\% on the HLD test set, a highly significant improvement over
chance, $p<.0001$.

\subsubsection{Feature analyses}
The all-features tree is complex and thus not shown, but feature
usage by feature type is summarized in Table~\ref{tab:featuse-clust-143}.

\begin{table}[H]
\begin{singlespace}
\caption{Feature Usage for Main Feature Types in Detection of Incomplete Utterances and Non-Incomplete Utterances}
\label{tab:featuse-clust-143}
\begin{center}
\begin{tabular}{|l|l|}
\hline
Feature  & Usage \\
Type     & \\
\hline
\hline
Duration &  0.557  \\
Energy &  0.182 \\
Enrate &  0.130 \\
F0 &  0.087  \\
Pause & 0.044 \\
\hline
\end{tabular}
\end{center}
\end{singlespace}
\end{table}

As indicated, the most-queried feature for this analysis is duration.
Not surprisingly, Incomplete Utterances are shorter overall than
complete ones; certainly they are by definition shorter than their
completed counterparts.  However, duration cannot completely
differentiate Incomplete from Non-Incomplete utterances, because
inherently short DAs (e.g., Backchannels, Agreements) are also present in
the data. For these cases, other features such as energy and
enrate play a role. 

\begin{figure}[H]
\begin{singlespace}
\begin{center}
\includebargraph{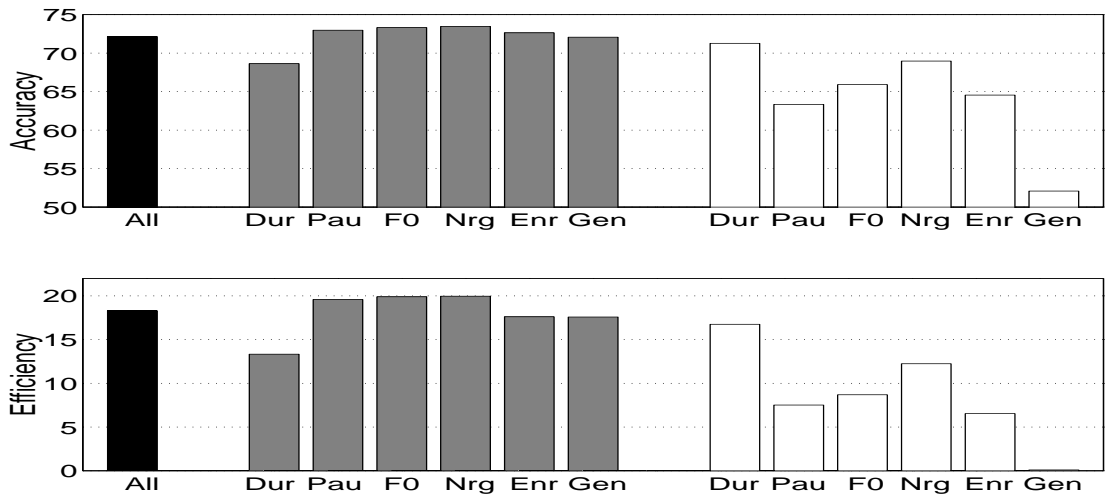}
\end{center}
\caption{Performance of prosodic trees using different feature sets
  for the detection of Incomplete Utterances from all other types.  $N$
  for each class=1323. Chance=50\%. Gray bars=exclude feature type;
  white bars=include only feature type.  Dur=Duration, Pau=Pause,
  F0=Fundamental frequency, Nrg=Energy, Enr=Speaking rate, Gen=Gender
  features.  }
\label{fig:bar143-Inc-Oth}
\end{singlespace}
\end{figure}

Results for trees run after features were selectively left out are shown
in Figure~\ref{fig:bar143-Inc-Oth}. Removal of duration features
resulted in a significant loss in accuracy, to 68.63\%, $p<.0001$. Removal
of any of the other feature types, however, did not significantly affect
performance.  Furthermore, a tree built using only duration features
yielded an accuracy of 71.28\%, which was not significantly less accurate
than the all-features tree. These results clearly indicate that duration
features are primary for this task. Nevertheless, good accuracy could be
achieved using other feature types alone; for all trees except the
gender-only tree, accuracy was significantly above chance, $p<.0001$.
Particularly noteworthy is the energy-only tree, which achieved an
accuracy of 68.97\%.  Typically, utterances fall to a low energy value
when close to completion. However, when speakers stop mid-stream, this
fall has not yet occurred, and thus the energy stays unusually high.
Inspection of the
energy-only tree revealed that over 75\% of the queries involved SNR
rather than RMS features, suggesting that at least for telephone speech,
it is crucial to use a feature
that can capture the energy from the speaker over the noise floor.

\subsubsection{Integration with language model}
We again integrated the all-features tree with a DA-specific language
model to determine whether prosody could aid classification with word
information present.  Results are presented in
Table~\ref{tab:integration-143}.  Like the earlier analyses, integration
improves performance over the words-only model for all three test cases.
Unlike earlier analyses, however, the relative improvement when true
words are used is minimal, and the effect is not significant for either
the HLD/true-words or the DEV/true-words data.  However, the relative improvement
for the DEV/N-best case is much larger.  The effect is just below
the significance threshold for this small dataset ($p=.067$), but would
be expected, based on the pattern of results in the previous analyses, to
easily reach significance for a set of data the size of the HLD set.

\begin{table}[H]
\begin{singlespace}
\caption{Accuracy of Individual and Combined Models for the Detection of Incomplete Utterances}
\label{tab:integration-143}
\begin{center}
\begin{tabular}{|r|c|c|c|}
\hline
Knowledge & HLD Set & DEV Set & DEV Set \\
Source    & true words  &  true words & N-best output \\         
\hline
\hline
 samples           &  2646   & 366   &  366   \\
 chance (\%)       & 50.00   & 50.00 & 50.00 \\ 
 tree   (\%)       & 72.16   & 72.01 & 72.01 \\
 words   (\%)      & 88.44   & 89.91 & 82.38 \\
 words+tree (\%)   & 88.74   & 90.49 & 84.56 \\
\hline
\end{tabular}
\end{center}
\end{singlespace}
\end{table}

Results suggest that for this task, prosody is an important knowledge
source when word recognition is not perfect. When true words are available,
however, it is not clear whether adding prosody aids performance. 
One factor underlying this pattern of results may be that the tree
information is already accounted for in the language model. Consistent
with this possibility is the fact that the tree uses mainly duration
features, which are indirectly represented in the language model by the
end-of-sentence marker. On the other hand, typically the word lengths of true and
N-best lists are similar, and our results differ for the two cases, so it is unlikely that 
this could be the only factor. 

Another possibility is that when true words are available, certain
canonical Incomplete Utterances can be detected with excellent accuracy.
A likely candidate here is the turn exit. Turn exits typically contain
one or two words from a small inventory of possibilities---mainly
coordinating conjunctions (``and'', ``but'') and fillers (``uh'', ``um'').
Similarly, because Switchboard consists mainly of first-person
narratives, a typical self-interrupted utterance in this corpus is a
noncommittal false start such as ``I---'' or ``I think---''. Both the turn
exits and the noncommittal false starts are lexically cued and are thus
likely to be well captured by a language model that has true words
available.

A third possible reason for the lack of improvement over true words is
that the prosodic model loses sensitivity because it averages over
phenomena with different characteristics.  False starts in our data
typically involved a sudden cut-off, whereas for turn exits the
preceding speech was often drawn out as in a hesitation.  As a
preliminary means of investigating this possibility, we built a tree for
Incomplete Utterances only, but breaking down the class into those
ending at turn boundaries (mainly turn exits and interrupted utterances)
versus those ending within a speaker's turn (mainly false starts).  The
resulting tree achieved high accuracy (81.17\%) and revealed that the
two subclasses differed on several features. For example, false
starts were longer in duration, higher in energy, and had faster
speaking rates than the turn exit/other-interrupted class.
Thus, as we also saw
for the case of Question detection, a prosodic model for Incomplete
Utterances is probably best built on data that have been broken down to
isolate subsets of phenomena whose prosodic features pattern
differently.

\subsection{Subtask 3: Detection of Agreements}

Our final subtask examined whether prosody could aid in the
detection of explicit Agreements (e.g., ``that's
exactly right'').  As shown earlier, Agreements were most often
misclassified as Backchannels (e.g., ``uh-huh'', ``yeah''). Thus, our
experiments focused on the distinction by including only these two DAs
in the trees.  An all-features tree for this task classified the 
data with an accuracy of 68.77\% (significantly above chance by a
binomial test, $p<.0001$) and with an efficiency of 12.21\%.

\subsubsection{Feature analyses}
\begin{figure}[H]
\begin{singlespace}
\begin{center}
\strut\psfig{file=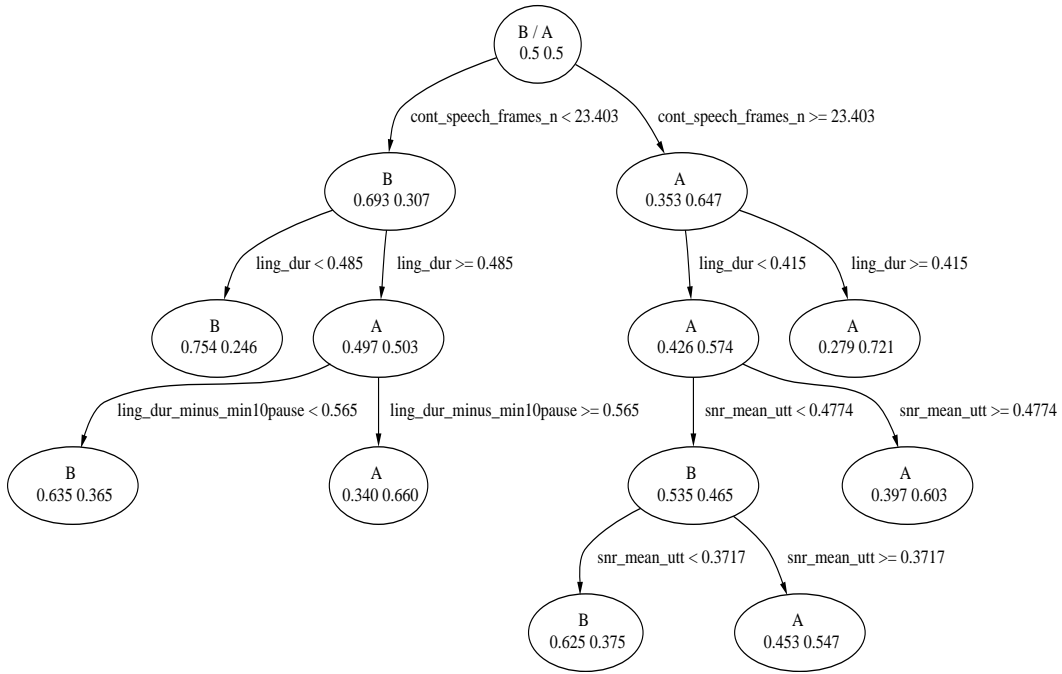,angle=270,width=5.5in,height=3.5in}
\end{center}
\caption{Decision tree for the classification of Backchannels (B) and Agreements (A)}
\label{fig:tree-145}
\end{singlespace}
\end{figure}

The all-features tree is shown in Figure~\ref{fig:tree-145}.  It uses
duration, pause, and energy features. From inspection we see that
Agreements are consistently longer in duration and have higher energy
(as measured by mean SNR) than Backchannels. The pause feature in this
case may play a role similar to that discussed for the question
classification task. Although Agreements and Backchannels were about
equally likely to occur turn-finally, Backchannels were more than three
times as likely as Agreements to be the {\em only} DA in a turn.  Thus,
Backchannels were more often surrounded by nonspeech regions (pauses
during which the other speaker was typically talking), causing the
cont\_speech\_frames window to not be filled at the edges of the DA and
thereby lowering the value of the feature.

Significance tests for the leave-one-out trees showed that removal of
the main feature types used in the all-features tree---that is, duration,
pause, and energy features---resulted in a significant reduction in
classification accuracy: $p<.001$, $p<.05$, and $p<.05$, respectively.
Although significant, the reduction was not large in absolute terms, as
seen from the figure and the $\alpha$ levels for significance.  For the
leave-one-in trees, results were in all cases significantly lower than
that of the all-features trees; however, duration and pause features
alone each yielded accuracy rates near that of the all-features tree.
Although neither F0 nor enrate was used in the all-features tree, each
individually was able to distinguish the DAs at rates significantly better
than chance ($p<.0001$).

\begin{figure}[H]
\begin{singlespace}
\begin{center}
\includebargraph{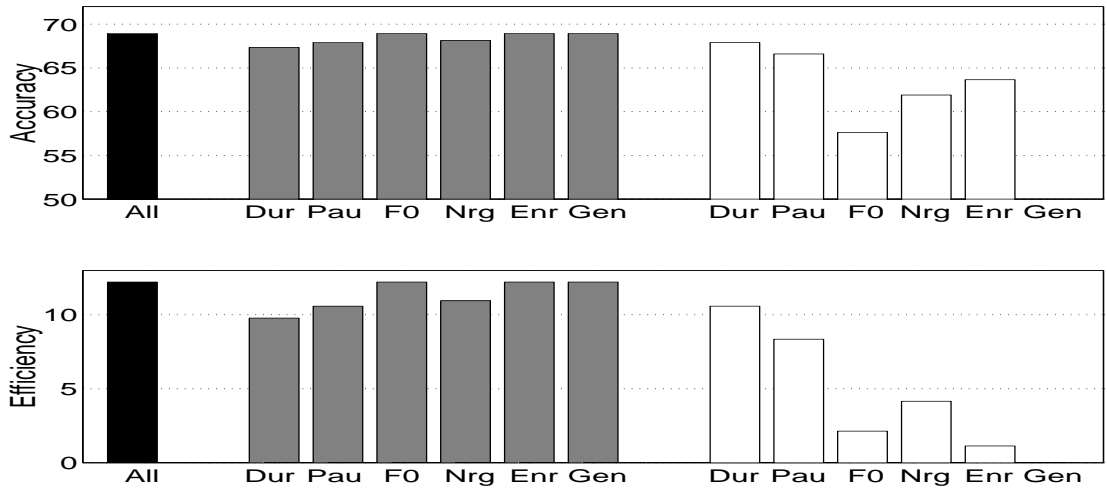}
\end{center}
\caption{Performance of prosodic trees using different feature sets
  for the classification of Backchannels and Agreements.  $N$ for each
  class=1260. Chance=50\%. Gray bars=exclude feature type; white
  bars=include only feature type. Dur=Duration, Pau=Pause,
  F0=Fundamental frequency, Nrg=Energy, Enr=Speaking rate, Gen=Gender
  features. }
\label{fig:bar145-Bac-Agr}
\end{singlespace}
\end{figure}

\subsubsection{Integration with language model}
Integration results are reported in Table~\ref{tab:integration-145}.
Several observations are noteworthy. First, integrating the tree with
word models improves performance considerably for all three test sets.
Sign tests run for the larger HLD set showed a highly significant gain
in accuracy by adding prosody, $p<.00001$. The DEV set did not contain
enough samples for sufficient power to reject the null hypothesis, but
showed the same pattern of results as the HLD set for both true and 
recognized words, and thus would be expected to reach
significance for a larger data set.  Second, for this analysis, the
prosodic tree has better accuracy than the true words for the HLD set.
Third, comparison of the data for the different test sets reveals an
unusual pattern of results. Typically (and in the previous analyses),
accuracy results for tree and word models were better for the HLD than
for the DEV set. As noted in the Method section, HLD waveforms were
segmented into DAs in the same manner (automatically) as the training
data, while DEV data were carefully segmented by hand. For this task, however,
results for both tree and word models are considerably better for the
DEV data, i.e., the mismatched case
(see also Figure~\ref{fig:lmconf}).
This pattern can be understood as
follows. In the automatically segmented training and HLD data,
utterances with ``bad'' estimated start or end times were thrown out of
the analysis, as described in the Method section. The DAs most affected
by the bad time marks were very short DAs, many of which were brief,
single-word Backchannels such as ``yeah''. Thus, the data remaining in
the training and HLD sets are biased toward longer DAs, while the data
in the DEV set retain the very brief DAs.  Since the present task pits
Backchannels against the longer Agreements, an increase in the
percentage of shorter Backchannels (from training to test, as occurs
when testing on the DEV data) can only enhance discriminability for
the prosodic trees as well as for the language model.  

\begin{table}[H]
\begin{singlespace}
\caption{Accuracy of Individual and Combined Models for the Detection of Agreements}
\label{tab:integration-145}
\begin{center}
\begin{tabular}{|r|c|c|c|}
\hline
Knowledge & HLD Set & DEV Set & DEV Set \\
Source    & true words  &  true words & N-best output \\         
\hline
\hline
 samples           & 2520  & 214    & 214  \\
 chance (\%)       & 50.00 & 50.00  & 50.00 \\ 
 tree   (\%)       & 68.77 & 72.88  & 72.88 \\
 words   (\%)      & 68.63 & 80.99  & 78.22  \\
 words+tree (\%)   & 76.90 & 84.74  & 81.70 \\
\hline
\end{tabular}
\end{center}
\end{singlespace}
\end{table}

\section{Summary and General Discussion}

\subsection{Feature Importance}

Across analyses we found that a variety of features were useful for DA
classification.  Results from the leave-one-out and leave-one-in trees
showed that there is considerable redundancy in the features; typically
there is little loss when one feature type is removed. Interestingly,
although canonical or predicted features such as F0 for questions are
important, less predictable features (such as pause features for
questions) show similar or even greater influence on results.

Duration was found to be important not only in the seven-way
classification, which included both long and short utterance types,
but also for subtasks within general length categories
(e.g., Statements versus Questions, Backchannels versus Agreements).
Duration was also found to be useful as an added knowledge source to
language model information, even though the length in words of an
utterance is indirectly captured by the language model.  Across tasks,
the most-queried duration features were not raw duration, but rather
duration-related measures that relied on the computation of other
feature types. 

F0 information was found to be important, as expected, for the
classification of Questions, particularly when questions were broken
down by type.  However, it was also of use in many other classification
tasks.  In general, the main contribution from F0 features for all but
the Question task came from global features (such as overall mean or
gradient) rather than local features (such as the penultimate and end
features, or the intonational event features).  An interesting issue
to explore in future work is whether this is a robustness effect, or
whether global features are inherently better predictors of DAs than
local features such as accents and boundaries.

Energy features were particularly helpful for classifying Incomplete
Utterances, but also for the classification of Agreements and
Backchannels.  Analysis of the usage of energy features over all tasks
revealed that SNR-based features were
queried more than 4.8 times as often as features based on the raw RMS
energy. Similarly, when the individual leave-one-in analyses for
energy features were computed using only RMS versus only SNR features,
results were consistently better for the SNR experiments.  This
suggests that for telephone speech or speech data collected under noisy
conditions, it is important to estimate the energy of the speaker
above the noise floor.

Enrate, the experimental speaking-rate feature from \namecite{MorganEtAl:97},
proved to be useful across analyses in the following way. Although no task was
significantly affected when enrate features were removed, enrate
systematically achieved good performance when used alone. It was
always better alone than at least one of the other main prosodic feature types
alone. Furthermore, it provided remarkable accuracy
for the classification of Questions and Statements, without any conversation-level 
normalization. Thus, the measure could be a valuable feature to include in a system,
particularly if other more costly features cannot be computed.

Finally, across analyses, gender was not used in the trees.  This
suggests that gender-dependent features such as F0 were sufficiently
normalized to allow gender-independent modeling.  Since many of the
features were normalized with respect to all values from a
conversation side, it is possible that men and women do differ in the
degree to which they use different prosodic features (even after
normalization for pitch range), but that we cannot discern these
differences here because speakers have been normalized individually.

Overall, the high degree of feature compensation found across tasks
suggests that automatic systems could be successful using only a
subset of the feature types.  However, we also found that
different feature types are used to varying degrees in the different
tasks, and it is not straightforward at this point to predict which
features will be most important for a task.  Therefore, for best
coverage on a variety of classification tasks, it is desirable to have
as many different feature types available as possible.

\subsection{Integration of Trees with Language Models}

Not only were the prosodic trees able to classify the data at rates
significantly above chance, but they also provided a consistent 
advantage over word information alone. To summarize the integration
experiments: all tasks with the exception of the Incomplete Utterance
task showed a significant improvement over words alone for the HLD
set.  For the Incomplete Utterance task, results for the DEV set were
marginally significant. In all cases, the DEV set lacked power because of
small sample size, making it difficult to reach significance in the
comparisons. However, the relative win on the DEV set was consistently
larger for the experiments using recognized rather than true
words. This pattern of results suggests that prosody can provide
significant benefit over word information alone, particularly when word
recognition is imperfect.

\section{Future Work}

\subsection{Improved DA Classification}

One aim of future work is to optimize the prosodic
features, and better understand the correlations among them. In
evaluating the contribution of features, it is important to take into
account such factors as measurement robustness and inherent
constraints leading to missing data in our trees.  For example,
duration is used frequently, but it is also (unlike, e.g., F0
information) available and fairly accurately extracted for all
utterances.  We would also like to better understand which of our
features capture functional versus semantic or paralinguistic
information, as well as the extent to which features are speaker-dependent.

A second goal is to explore additional features that do not depend on
the words.  For example, we found that whether or not an utterance is
turn-initial and/or turn-final, and the rate of interruption
(including overlaps) by the other speaker, can significantly improve
tree performance for certain tasks.  In our overall model, we consider
turn-related features to be part of the dialog grammar.  Nevertheless,
if one wanted to design a system that did not use word information,
turn features could be used along with the prosodic features to
improve performance overall.

Third, although we chose to use decision trees for the reasons given
earlier, we might have used any suitable probabilistic classifier,
i.e., any model that estimates the posterior probabilities of DAs
given the prosodic features. We have conducted preliminary experiments
to assess how neural networks compare to decision trees for the type
of data studied here.  Neural networks are worth investigating since
they offer potential advantages over decision trees.  They
can learn decision surfaces that lie at an angle to the axes of the
input feature space, unlike standard CART trees, which always split
continuous features on one dimension at a time.  The response function
of neural networks is continuous (smooth) at the decision boundaries,
allowing them to avoid hard decisions and the complete fragmentation
of data associated with decision tree questions.  Most
important, neural networks with hidden units can learn new features
that combine multiple input features.  Results from preliminary
experiments on a single task showed that a softmax network
\cite{Bridle:90} without hidden units resulted in a slight improvement
over a decision tree on the same task. The fact that hidden units did
not afford an advantage indicates that complex combinations of
features (as far as the network could learn them) may not better
predict DAs for the task than linear combinations of our input
features.

Thus, whether or not substantial gains can be obtained using
alternative classifier architectures remains an open question.
One approach that looks promising given the redundancy
among different feature types is a combination of parallel
classifiers, each based on a subcategory of features, for example using the
mixture-of-experts framework \cite{Jordan:94}.  We will also need to
develop an effective way to combine specialized classifiers (such as
those investigated for the subtasks in this study) into an overall
classifier for the entire DA set.

Finally, many questions remain concerning the best way to integrate
the various knowledge sources.  Instead of treating words and prosody
as independent knowledge sources, as done here for simplicity, we
could provide both types of cues to a single classifier.  This would
allow the model to account for interactions between prosodic cues and
words, such as word-specific prosodic patterns.  The main problem with such
an approach is
the large number of potential input values that ``word features'' can
take on.
A related question is how to combine prosodic classifiers
most effectively with dialog grammars and the contextual knowledge
sources.

\subsection{Automatic Dialog Act Classification and Segmentation}

Perhaps the most important area for future work is the automatic
segmentation of dialogs into utterance units.
As explained earlier, we side-stepped the segmentation problem for
the present study by using segmentations
by human labelers.  Eventually, however, a fully automatic dialog
annotation system will have to perform both segmentation and DA classification.
Not only is this combined task more difficult, it also raises
methodological issues,
such as how to evaluate the DA classification on incorrectly identified
utterance units.  One approach, taken by \namecite{MastEtAl:96}, is to 
evaluate recognized DA sequences in terms of substitution, deletion, and
insertion errors, analogous to the scoring of speech recognition output.

As noted in the Introduction, a large body of work addresses
segmentation into intonational units or prosodic phrases, and utterance
segmentation can be considered as a special case of prosodic boundary
detection. To our knowledge, there are no published results for
performing utterance-level segmentation of spontaneous speech by using only
acoustic evidence, i.e., without knowledge of the correct words.  
Studies have investigated segmentation assuming that some kind
of word-level information is given.  \namecite{MastEtAl:96} and
\namecite{Warnke:97} investigate DA segmentation and classification in
the (task-oriented) Verbmobil domain, combining neural-network prosodic
models with N-gram models for segment boundary detection, as well as
N-gram and decision tree DA models with N-gram discourse grammars for DA
classification, in a mathematical framework very similar to the one used
here.  \namecite{StoShr:icslp96} and \namecite{FinkeEtAl:aaai98} both
investigated segmentation of spontaneous, Switchboard-style
conversations using word-level N-gram models.  \namecite{StoShr:icslp96}
observed that word-level N-gram segmentation models work best when
using a combination of parts-of-speech and cue words, rather than words
alone.

Both \namecite{Warnke:97} and \namecite{FinkeEtAl:aaai98} propose an 
A$^\ast$ search for integrated DA segmentation and labeling.
However, the results of \namecite{Warnke:97} show only a small improvement
over a sequential (first segment, then label) approach, and
\namecite{FinkeEtAl:aaai98} found that segmentation accuracy did not
change significantly as a result of modeling DAs in the segment language model.
These findings indicate that a DA-independent utterance segmentation, followed
by DA labeling using the methods described here, will be a reasonable 
strategy for extending our approach to unsegmented speech.
This is especially important since our prosodic features rely on known 
utterance boundaries for extraction and normalization.

\subsection{Dialog Act Classification and Word Recognition}

As mentioned in the Introduction, in addition to dialog modeling
as a final goal, there are other practical
reasons for developing methods for automatic DA classification.
In particular, DA classification holds
the potential to improve speech recognition accuracy, since 
language models constrained by the DA can be applied when the utterance
type is known.
There has been little work involving speech recognition output for large
annotated natural speech corpora.  One relevant experiment
has been conducted as part of our larger
WS97 discourse modeling project, described in detail elsewhere
\cite{JuretalTR:98}.

To put an upper bound on the potential benefit of the approach, it is
most meaningful to consider the extent to which word recognition
accuracy could be improved if one's automatic DA classifier had perfect
accuracy. We therefore conducted experiments in which our language
models were conditioned on the correct (i.e., hand-labeled) DA type.
From the perspective of overall word accuracy results, the outcome was
somewhat discouraging.  Overall, the word error rate dropped by only
0.9\% absolute, from a baseline of 41.2\% to 40.9\%. On the other hand,
if one considers the Switchboard corpus statistics, results are in line
with what one would predict for this corpus.  In Switchboard, roughly
83\% of all test set words were contained in the Statement category.
Statements are thus already well-represented in the baseline language
model. It is not surprising, then, that the error rate for Statements
was reduced by only 0.5\%.  The approach was successful, however, for
reducing word error for other DA types. For example, for Backchannels
and No-Answers, word error was reduced significantly (by 7\% and 18\%,
respectively). But because these syntactically restricted categories
tend to be both less frequent and shorter than Statements, they
contributed too few words to have much of an impact on the overall word
error rate.

The DA-specific error reduction results suggest that although overall
word accuracy for Switchboard was little improved in our experiments, DA
classification could substantially benefit word recognition results for
other types of speech data, or when evaluating on specific DA types.
This should be true particularly for domains with a less skewed
distribution of DA types.  Similarly, DA modeling could reduce word
error for corpora with a more uniform distribution of utterance lengths,
or for applications where it is important to correctly recognize words
in a specific subset of DAs.

\section{Conclusion}

We have shown that in a large database of natural human-human
conversations, assuming equal class prior probabilities, prosody is a
useful knowledge source for a variety of DA classification tasks.  The
features that allow this classification are task-dependent.  Although
canonical features are used in predicted ways, other less obvious
features also play important roles. Overall there is a high degree of
correlation among features, such that if one feature type is not
available, other features can compensate. Finally, integrating prosodic
decision trees with DA-specific statistical language models improves
performance over that of the language models alone, particularly
in a realistic setting where word information is based on automatic recognition.
We conclude that
DAs are redundantly marked in free conversation, and that a variety of
automatically extractable prosodic features could aid the processing of
natural dialog in speech applications.

\newpage
\bibliographystyle{myapa}
\bibliography{strings,draft,dan,proc}

\newpage

\appendix

\section{Appendix A: Table of Original Dialog Acts}

The following table lists the 42 original (before grouping into classes)
dialog acts.
Counts and relative frequencies were obtained from the corpus
of 197,000 utterances used in model training.

\begin{singlespace}
\begin{tabular}{|l|c|l|r|r|}
\hline
{\bf Dialog Act} & {\bf Tag} & {\bf Example} & {\bf Count} & {\bf \%} \\
\hline
\hline
Statement-non-opinion&sd &{\em  Me,  I'm in the legal department.} &72,824 & 36\\
Acknowledge (Backchannel) &b& {\em Uh-huh.} & 37,096 & 19\\
Statement-opinion&sv&  {\em I think it's great.}& 25,197 & 13\\
Agree/Accept&aa& {\em That's exactly it.} & 10,820 & 5\\
Abandoned or Turn-Exit &\% \ldots -/&{\em So, -/}&10,569 & 5 \\
Appreciation&ba&{\em I can imagine.}& 4,633 & 2\\
Yes-No-Question&qy&{\em Do you have to have any special training?}&4,624 & 2\\
Non-verbal & x & {\em $<$Laughter$>$,$<$Throat\_clearing$>$} &3,548 & 2 \\
Yes-Answer &ny & {\em Yes.} & 2,934 & 1\\
Conventional-closing&fc&{\em Well, it's been nice talking to you.}& 2,486 &1\\
Uninterpretable& \%&{\em But, uh, yeah. }&2,158 & 1\\
Wh-Question& qw&{\em Well,  how old are you?}&1,911 & 1\\
No-Answer&nn& {\em No.} & 1,340 & 1\\
Acknowledge-Answer&bk&{\em Oh, okay.}&1,277& 1 \\
Hedge&h& {\em I don't know if I'm making any sense or not.} &1,182 & 1 \\
Declarative Yes-No-Question & qy\^{}d&{\em So you can afford to get a house?}& 1,174 & 1\\
Other&o,fo& {\em  Well give me a break, you know. }&1,074 & 1\\
Backchannel-Question &bh& {\em Is that right?} & 1,019 & 1\\
Quotation&\^{}q& {\em He's always saying ``why do they have to be here?''} & 934&.5\\
Summarize/Reformulate&bf& {\em Oh, you mean you switched schools for the kids.} & 919 & .5\\
Affirmative Non-Yes Answers&na & {\em It is.} & 836 &.4\\
Action-directive          & ad& {\em Why don't you go first}& 719 & .4\\
Collaborative Completion& \^{}2&{\em Who aren't contributing.}&699&.4\\
Repeat-phrase&b\^{}m& {\em Oh, fajitas.} & 660 & .3\\
Open-Question& qo&{\em How about you?}&632 & .3\\
Rhetorical-Questions& qh&{\em Who would steal a newspaper?}&557&.2\\
Hold before Answer/Agreement&\^{}h& {\em I'm drawing a blank.} & 540 & .3\\
Reject&ar& {\em Well, no.} &338&.2 \\
Negative Non-No Answers&ng & {\em  Uh, not a whole lot.} &292 & .1 \\
Signal-non-understanding&br& {\em Excuse me?} & 288 & .1\\
Other Answers&no   & {\em I don't know.} & 279 & .1\\
Conventional-opening&fp&{\em How are you?}&220 & .1\\
Or-Clause& qrr&{\em or is it more of a company?} &207 & .1\\
Dispreferred Answers&arp,nd &{\em Well, not so much that.}& 205 & .1\\
Third-party-talk    & t3& {\em My goodness, Diane, get down from there.} & 115 &.1\\
Offers, Options \& Commits&oo,cc,co& {\em I'll have to check that out.} &109 & .1\\
Self-talk&t1  &{\em What's the word I'm looking for?} & 102 & .1\\
Downplayer&bd& {\em That's all right.} & 100 & .1\\
Maybe/Accept-part&aap/am& {\em Something like that.}& 98 & $<$.1\\
Tag-Question& \^{}\ g&{\em Right?} &93 & $<$.1\\
Declarative Wh-Question& qw\^{}d&{\em You are what kind of buff?}&80 & $<$.1\\
Apology&fa& {\em I'm sorry.} &76 & $<$.1 \\
Thanking&ft& {\em  Hey thanks a lot.} &67 & $<$.1\\
\hline
\end{tabular}
\end{singlespace}
\newpage

\section{Appendix B: Estimated accuracy of transcript-based labeling}

The table below shows the estimated recall and precision of 
hand-labeling utterances using only the transcribed words.

The estimates are computed using the results of ``Round I'' relabeling
with listening to speech (see the Method section) as reference labels.
DA types are sorted by their occurrence count in the relabeled subcorpus
of 44 conversations.

For a given DA type, let $a$ be the number of original (labeled
from text only) DA tokens of that type, $b$ the number of DA tokens
after relabeling with listening, and $c$ the number of tokens that
remained unchanged in the relabeling.  Recall is estimated as $b
\over a$ and precision as $c \over a$.

\begin{center}
\begin{singlespace}
\begin{tabular}{|l|c|r|r|r|}
\hline
{\bf Dialog Act} & {\bf Tag} & {\bf Recall (\%)} & {\bf Precision (\%)} & {\bf Count} \\
\hline
\hline
 Statement-non-opinion & sd & 98.8 & 98.9 & 2147 \\
 Statement-opinion & sv & 97.9 & 97.7 & 989 \\
 Acknowledge (Backchannel) & b & 99.1 & 95.4 & 986 \\
 Abandoned/Uninterpretable & \% & 99.8 & 99.4 & 466 \\
 Agree/Accept & aa & 86.5 & 99.3 & 327 \\
 Yes-No-Question & qy & 100.0 & 98.0 & 144 \\
 Non-verbal & x & 100.0 & 100.0 & 99 \\
 Appreciation & ba & 100.0 & 94.6 & 70 \\
 Yes-Answer & ny & 95.7 & 98.5 & 70 \\
 Wh-Question & qw & 98.3 & 100.0 & 59 \\
 Summarize/Reformulate & bf & 100.0 & 97.8 & 44 \\
 Hedge & h & 93.0 & 97.6 & 43 \\
 Quotation & \^{}q & 100.0 & 100.0 & 38 \\
 Declarative Yes-No-Question & qy\^{}d & 92.1 & 97.2 & 38 \\
 Acknowledge-Answer & bk & 100.0 & 100.0 & 34 \\
 No-Answer & nn & 100.0 & 100.0 & 33 \\
 Other & o,fo & 100.0 & 100.0 & 33 \\
 Open-Question & qo & 100.0 & 100.0 & 27 \\
 Backchannel-Question & bh & 95.5 & 100.0 & 22 \\
 Action-directive & ad & 100.0 & 95.5 & 21 \\
 Collaborative Completion & \^{}2 & 100.0 & 94.7 & 18 \\
 Hold before Answer/Agreement & \^{}h & 100.0 & 100.0 & 18 \\
 Affirmative Non-Yes Answers & na & 100.0 & 100.0 & 18 \\
 Repeat-phrase & b\^{}m & 100.0 & 100.0 & 17 \\
 Conventional-closing & fc & 100.0 & 100.0 & 16 \\
 Reject & ar & 100.0 & 100.0 & 13 \\
 Or-Clause & qrr & 100.0 & 100.0 & 11 \\
 Other Answers & no & 100.0 & 100.0 & 10 \\
 Rhetorical-Questions & qh & 80.0 & 100.0 & 10 \\
 Signal-non-understanding & br & 100.0 & 87.5 & 7 \\
 Negative Non-No Answers & ng & 100.0 & 100.0 & 6 \\
 Maybe/Accept-part & aap/am & 100.0 & 100.0 & 5 \\
 Conventional-opening & fp & 100.0 & 100.0 & 5 \\
 Tag-Question & \^{}g & 100.0 & 100.0 & 4 \\
 Offers, Options \& Commits & oo,cc,co & 100.0 & 100.0 & 3 \\
 Thanking & ft & 100.0 & 100.0 & 2 \\
 Downplayer & bd & 100.0 & 100.0 & 1 \\
 Declarative Wh-Question & qw\^{}d & 100.0 & 100.0 & 1 \\
 Self-talk & t1 & 100.0 & 50.0 & 1 \\
 Third-party-talk & t3 & 100.0 & 100.0 & 1 \\
 Dispreferred Answers & arp,nd & - & - & 0 \\
 Apology & fa & - & - & 0 \\
\hline
\end{tabular}
\end{singlespace}
\end{center}
  
\end{document}